\documentclass{article}
\usepackage{arxiv}
\usepackage{graphicx}
\usepackage[numbers]{natbib}
\usepackage{doi}
\usepackage[utf8]{inputenc} %
\usepackage[T1]{fontenc}    %
\usepackage{url}            %
\usepackage{booktabs}       %
\usepackage{amsfonts}       %
\usepackage{nicefrac}       %
\usepackage{microtype}      %
\usepackage[table,dvipsnames]{xcolor}         %
\usepackage{amsmath}
\usepackage{amssymb}
\usepackage{bbm}
\usepackage{wasysym}
\usepackage{pifont}
\usepackage{threeparttable}
\usepackage{subcaption}
\usepackage{array} %
\usepackage{makecell}
\usepackage{algpseudocode} 
\usepackage{multirow}
\usepackage{listings}
\usepackage{enumitem}
\usepackage{hyperref}
\usepackage{todonotes}
\usepackage{comment}
\usepackage[capitalise, noabbrev]{cleveref}
\usepackage{lscape}
\usepackage{tabularray}
\usepackage{CJK}

\usepackage{wrapfig}
\usepackage{float}
\usepackage[ruled,vlined]{algorithm2e}

\usepackage{arydshln}
\usepackage{booktabs}
\usepackage{multirow}
\usepackage{graphicx}
\usepackage{pifont}
\usepackage{authblk}

\definecolor{taborange}{rgb}{1, 0.85, 0.7}
\definecolor{tabgreen}{rgb}{0.72, 0.9, 0.80}
\definecolor{tabgrey}{rgb}{0.9, 0.9, 0.9}

\usepackage{anyfontsize}

\newcommand{\cmark}{\textcolor{ForestGreen}{\ding{51}}} 
\newcommand{\xmark}{\textcolor{red!80!black}{\ding{55}}} %

\newcommand{\etal}{\textit{et al}. }

\newcommand{\eg}{\textit{e.g.}}

\newcommand{\ie}{\emph{i.e.}}

\usepackage{authblk}

\title{Spatial4D-Bench: A Versatile 4D Spatial Intelligence Benchmark}

\renewcommand\Affilfont{\normalsize} 
\makeatletter
\renewcommand\AB@affilsepx{, \protect\Affilfont}
\makeatother

\author[1,*]{Pan Wang}
\author[2,*]{Yang Liu}
\author[2]{Guile Wu}
\author[2]{Eduardo R. Corral-Soto}
\author[2]{Chengjie Huang}
\author[2]{Binbin Xu}
\author[2]{Dongfeng Bai}
\author[1]{Xu Yan}
\author[2]{Yuan Ren}
\author[2]{Xingxin Chen}
\author[1]{Yizhe Wu}
\author[1]{Tao Huang}
\author[3]{Wenjun Wan}
\author[3]{Xin Wu}
\author[3]{Pei Zhou}
\author[3]{Xuyang Dai}
\author[1,7]{Kangbo Lv}
\author[1]{Hongbo Zhang}
\author[1]{Yosef Fried}
\author[1]{Aixue Ye}
\author[1]{Bailan Feng}
\author[1]{Zhenyu Chen}
\author[4]{Zhen Li}
\author[5]{Yingcong Chen}
\author[6]{Yiyi Liao}
\author[1,$\dagger$]{Bingbing Liu}

\affil[1]{Fundation Model Dept, Huawei}
\affil[2]{Noah’s Ark Lab, Huawei}
\affil[3]{Central Media Technology Institute, Huawei}
\affil[4]{CUHK-Shenzhen}
\affil[5]{HKUST-GZ}
\affil[6]{Zhejiang University}
\affil[7]{Tsinghua University}

\renewcommand{\headeright}{Technical Report}

\begin{document}

\maketitle

\begingroup
\renewcommand\thefootnote{*}
\footnotetext{Equal contribution.\;$^\dagger$Corresponding author, liu.bingbing@huawei.com}
\endgroup

\thispagestyle{fancy}

\begin{abstract}
4D spatial intelligence involves perceiving and processing how objects move or change over time. Humans naturally possess 4D spatial intelligence, supporting a broad spectrum of spatial reasoning abilities.
To what extent can Multimodal Large Language Models (MLLMs) achieve human-level 4D spatial intelligence?
In this work, we present Spatial4D-Bench, a versatile 4D spatial intelligence benchmark designed to comprehensively assess the 4D spatial reasoning abilities of MLLMs.
Unlike existing spatial intelligence benchmarks that are often small-scale or limited in diversity, Spatial4D-Bench provides a large-scale, multi-task evaluation benchmark consisting of \textasciitilde40,000 question-answer pairs covering 18 well-defined tasks. We systematically organize these tasks into six cognitive categories: object understanding, scene understanding, spatial relationship understanding, spatiotemporal relationship understanding, spatial reasoning and spatiotemporal reasoning. Spatial4D-Bench thereby offers a structured and comprehensive benchmark for evaluating the spatial cognition abilities of MLLMs, covering a broad spectrum of tasks that parallel the versatility of human spatial intelligence.
We benchmark various state-of-the-art open-source and proprietary MLLMs on Spatial4D-Bench and reveal their substantial limitations in a wide variety of 4D spatial reasoning aspects, such as route plan, action recognition, and physical plausibility reasoning. We hope that the findings provided in this work offer valuable insights to the community and that our benchmark can facilitate the development of more capable MLLMs toward human-level 4D spatial intelligence. More resources can be found on our project page: \url{https://spatial4d-bench.github.io/spatial4d/}.
\end{abstract}
    
\section{Introduction}
\label{sec:intro}

In cognitive science, spatial cognition seeks to understand how humans and animals perceive, interpret, mentally represent, and interact with the spatial characteristics of the environment~\cite{waller2013handbook, newcombe2004spatial}.
To assess human spatial intelligence, a wide range of standardized tests have been developed over the past decades, such as puzzles, pattern blocks, tangrams, paper-and-pencil tests~\cite{carroll1993human,eliot1983international,hegarty2005individual}, Mental Rotations Test (MRT)~\cite{peters1995redrawn}, and navigation tests based on virtual 3D reality environments~\cite{wolbers2008spatial}.
Analogously, Multimodal Large Language Models (MLLMs) have been developed to unify language and vision, with the ultimate goal of achieving human-level spatial understanding and reasoning.
Recent developments in MLLMs~\cite{OpenAI2025,Gemini2025,zhang2025videollama,bai2025qwen2,wang2025internvl3} have achieved impressive performance, covering a wide range of multimodal understanding and reasoning tasks.
However, to what extent MLLMs can achieve human spatial cognition levels remains an open question.

To investigate this, researchers have recently developed some spatial intelligence benchmarks~\cite{Yang_2025_CVPR,yu2025far,liu2025mirage,gong2025space,deng2025internspatial,yang2025mmsi,ma20243dsrbench,cheng2024spatialrgpt,zhao2025mmvu,tang2025lego} and test a variety of MLLMs to assess their spatial reasoning capabilities.
These benchmarks predominantly assess foundational spatial reasoning capabilities, such as room size estimation, object distance reasoning, and object counting.
Although these benchmarks have revealed substantial capability limitations of existing MLLMs on spatial reasoning tasks, they are often small-scale and lack diversity.
This poses a challenge for comprehensively evaluating the spatial intelligence capabilities of MLLMs.
Moreover, most existing spatial intelligence benchmarks focus only on 3D spatial intelligence with a primary focus on reasoning about static scenes, largely neglecting the evaluation of spatiotemporal awareness.
Yet the real world is inherently a 4D environment, where spatial and temporal aspects continuously evolve.
Living in such a time-evolving 4D environment, humans naturally develop 4D spatial intelligence in which perceiving and processing how objects move or change dynamically over time are the key tasks, enabling a variety of spatial reasoning abilities.
In light of this, several spatial intelligence benchmarks have involved spatiotemporal reasoning, such as STI-Bench~\cite{li2025sti}, VSI-SUPER~\cite{yang2025cambrian}, and VLM4D~\cite{zhou2025vlm4d}.
However, the evaluation data for these benchmarks remain limited in terms of diversity or scale, limiting their ability to comprehensively evaluate MLLMs in 4D spatial reasoning and to assess the gap between MLLMs and human-level 4D spatial intelligence.

\begin{figure} 
	\centering	
		\includegraphics[width=0.99\linewidth]{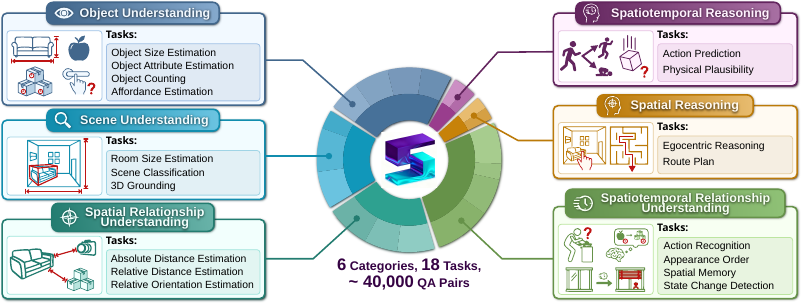} 		
	\caption{
    An overview of Spatial4D-Bench.
    Spatial4D-Bench is a large-scale, multi-task evaluation benchmark to comprehensively assess MLLMs' 4D spatial reasoning abilities.
    It consists of \textasciitilde40,000 question-answer pairs covering 18 well-defined tasks, which are organized into 6 categories, including object understanding, scene understanding, spatial relationship understanding, spatiotemporal relationship understanding, spatial reasoning and spatiotemporal reasoning, covering various aspects of 4D spatial reasoning.
    Example question-answer pairs can be found in the appendix.
	\label{fig_spatial4dbench_overview} }
\end{figure}

\begin{table}[t]
\centering
\footnotesize
\setlength{\tabcolsep}{2.5pt} 
\begin{tabular}{llcccccccc}
\toprule
 & & & & \multicolumn{6}{c}{\textbf{Cognitive Categories}} \\
 \cmidrule(lr){5-10}
 & & \textbf{\# Eval} & \textbf{\# } & 
\multicolumn{1}{c}{\textbf{Object}} & 
\multicolumn{1}{c}{\textbf{Scene}} & 
\multicolumn{1}{c}{\textbf{Spatial}} & 
\multicolumn{1}{c}{\textbf{S.T.}} & 
\multicolumn{1}{c}{\textbf{Spatial}} & 
\multicolumn{1}{c}{\textbf{S.T.}} \\ 
\textbf{Benchmark} & \textbf{Pub.} & \textbf{QA} & \textbf{Tasks} & 
\multicolumn{1}{c}{\textbf{Und.}} & 
\multicolumn{1}{c}{\textbf{Und.}} & 
\multicolumn{1}{c}{\textbf{Rel.}} & 
\multicolumn{1}{c}{\textbf{Rel.}} & 
\multicolumn{1}{c}{\textbf{Reas.}} & 
\multicolumn{1}{c}{\textbf{Reas.}} \\ 
\midrule

SpatialRGPT~\cite{cheng2024spatialrgpt} & Jun. 2024 & 1,406 & 2 & \xmark & \xmark & \cmark & \xmark & \xmark & \xmark \\

3DSRBench~\cite{ma20243dsrbench} & Dec. 2024 & 2,772 & 4 & \textcolor{Goldenrod}{\textbf{\LEFTcircle}} & \xmark & \cmark & \xmark & \xmark & \xmark \\

VSI-Bench~\cite{Yang_2025_CVPR} & Dec. 2024 & 5,156 & 8 & \textcolor{Goldenrod}{\textbf{\LEFTcircle}} & \textcolor{Goldenrod}{\textbf{\LEFTcircle}} & \cmark & \textcolor{Goldenrod}{\textbf{\LEFTcircle}} & \textcolor{Goldenrod}{\textbf{\LEFTcircle}} & \xmark \\

STI-Bench~\cite{li2025sti} & Mar. 2025 & 2,064 & 8 & \textcolor{Goldenrod}{\textbf{\LEFTcircle}} & \textcolor{Goldenrod}{\textbf{\LEFTcircle}} & \cmark & \textcolor{Goldenrod}{\textbf{\LEFTcircle}} & \xmark & \xmark \\

LEGO-Puzzles~\cite{tang2025lego} & Mar. 2025 & 1,100 & 11 & \xmark & \xmark & \xmark & \textcolor{Goldenrod}{\textbf{\LEFTcircle}} & \xmark & \textcolor{Goldenrod}{\textbf{\LEFTcircle}} \\

MMSI-Bench~\cite{yang2025mmsi} & May. 2025 & 1,000 & 4 & \xmark & \xmark & \cmark & \textcolor{Goldenrod}{\textbf{\LEFTcircle}} & \xmark & \xmark \\

MIRAGE~\cite{liu2025mirage} & May. 2025 & 1,710 & 3 & \textcolor{Goldenrod}{\textbf{\LEFTcircle}} & \xmark & \textcolor{Goldenrod}{\textbf{\LEFTcircle}} & \textcolor{Goldenrod}{\textbf{\LEFTcircle}} & \xmark & \xmark \\

SpaCE-10~\cite{gong2025space} & Jun. 2025 & 5,000 & 8 & \textcolor{Goldenrod}{\textbf{\LEFTcircle}} & \textcolor{Goldenrod}{\textbf{\LEFTcircle}} & \cmark & \xmark & \xmark & \xmark \\

InternSpatial~\cite{deng2025internspatial} & Jun. 2025 & 6,008 & 4 & \textcolor{Goldenrod}{\textbf{\LEFTcircle}} & \xmark & \cmark & \xmark & \xmark & \xmark \\

VLM4D~\cite{zhou2025vlm4d} & Aug. 2025 & 1,816 & 4 & \xmark & \xmark & \xmark & \cmark & \xmark & \xmark \\

SIBench~\cite{yu2025far} & Sep. 2025 & 9,000 & 7 & \textcolor{Goldenrod}{\textbf{\LEFTcircle}} & \textcolor{Goldenrod}{\textbf{\LEFTcircle}} & \cmark & \textcolor{Goldenrod}{\textbf{\LEFTcircle}} & \textcolor{Goldenrod}{\textbf{\LEFTcircle}} & \textcolor{Goldenrod}{\textbf{\LEFTcircle}} \\

\midrule
\textbf{Spatial4D-Bench} & \textbf{Ours} & \textbf{39,305} & \textbf{18} & \cmark & \cmark & \cmark & \cmark & \cmark & \cmark \\ 
\midrule[0.08em]
\multicolumn{10}{c}{\cmark: fully covered\quad  \textcolor{Goldenrod}{\textbf{\LEFTcircle}}: partially covered \quad \xmark: not covered}
\end{tabular}%
\vspace{0.5em}
    \caption{Comparison of Spatial4D-Bench with state-of-the-art spatial intelligence benchmarks. We evaluate coverage across 6 cognitive categories: object understanding (size, attribute, count and affordance), scene understanding (room size, scene class and grounding), spatial relationships (absolute/relative distance and orientation), spatiotemporal (S.T.) relationships (action, order, memory and state change), spatial reasoning (egocentric and route plan), and spatiotemporal reasoning (prediction and physical plausibility). Unlike prior works, Spatial4D-Bench provides significantly higher data scale and comprehensive coverage of all 18 tasks, offering a robust evaluation of MLLMs' 4D reasoning capabilities.}
    \label{tab:recent_mllm_eval_benchmarks}
\end{table}

In this work, we present \textbf{Spatial4D-Bench}, a large-scale, multi-task 4D spatial intelligence benchmark that enables comprehensive assessment of MLLMs' spatial reasoning abilities. As shown in \cref{fig_spatial4dbench_overview}, Spatial4D-Bench comprises \textasciitilde40,000 carefully curated and annotated question-answer (QA) pairs, covering a wide variety of indoor and outdoor environments involving diverse objects, actions, and scenes.
By adhering to human spatial cognition principles~\cite{waller2013handbook, newcombe2004spatial}, these QA pairs are divided into 6 categories, namely, \textit{object understanding}, \textit{scene understanding}, \textit{spatial relationship understanding}, \textit{spatiotemporal relationship understanding}, \textit{spatial reasoning}, and \textit{spatiotemporal reasoning}.
Each category is further subdivided into various tasks, yielding 18 tasks in total that span a broad range of spatial perception, understanding and reasoning abilities.
This significantly distinguishes Spatial4D-Bench from existing benchmarks that are often small-scale or limited in diversity.
Although existing benchmarks have covered some tasks (\eg, object size estimation and object counting) overlapping with Spatial4D-Bench, some 4D tasks presented in Spatial4D-Bench remain insufficiently investigated, including but not limited to \textit{spatial memory}, \textit{state change detection}, and \textit{physical plausibility reasoning}.
Therefore, compared to existing benchmarks, Spatial4D-Bench provides a more comprehensive evaluation suite for the assessment of MLLMs' spatial cognition abilities, spanning a variety of tasks that parallel the versatility of human spatial intelligence. 
\cref{tab:recent_mllm_eval_benchmarks} summarizes the statistical differences between Spatial4D-Bench and existing benchmarks.

We conduct thorough experiments to benchmark a variety of state-of-the-art MLLMs on Spatial4D-Bench, including two proprietary MLLMs (GPT-5~\cite{OpenAI2025} and Gemini 2.5-Pro~\cite{Gemini2025}) and several open-source MLLMs (VideoLLama3~\cite{zhang2025videollama}, Qwen2.5-VL~\cite{bai2025qwen2}, Qwen3-VL~\cite{bai2025qwen2}, and InternVL3.5~\cite{wang2025internvl3}) with model sizes ranging from 7B to 241B parameters.
Our extensive experiments and in-depth analysis reveal that MLLMs still exhibit a performance gap relative to humans in comprehensive 4D spatial reasoning.
In particular, MLLMs have substantial limitations in a wide variety of 4D spatial reasoning aspects, such as route plan, egocentric reasoning, and physical plausibility reasoning.
Nevertheless, we also observe that MLLMs surpass human performance on certain tasks, such as room size and object size estimation.
This is reasonable, as humans generally struggle with tasks that require the accurate estimation of 3D scale in real world, while MLLMs can outperform humans by leveraging vast amounts of training data to provide prior knowledge.
We hope that these findings provide valuable insights to the community and that the release of Spatial4D-Bench facilitates the development of more capable MLLMs toward human-level 4D spatial intelligence.
More resources can be found on our project page: \url{https://spatial4d-bench.github.io/spatial4d/}.

\section{Related Work}
\label{sec:related_works}

\subsection{Spatial Intelligence Benchmark}
Multiple spatial intelligence benchmarks~\cite{Yang_2025_CVPR,yu2025far,liu2025mirage,gong2025space,deng2025internspatial,yang2025mmsi,ma20243dsrbench,cheng2024spatialrgpt,zhao2025mmvu,tang2025lego} have recently emerged to evaluate MLLM's capabilities in spatial reasoning tasks.
VSI-Bench utilizes~\cite{Yang_2025_CVPR} public 3D scene datasets including ScanNet~\cite{dai2017scannet}, ScanNet++~\cite{yeshwanth2023scannet++} and ARKitScenes~\cite{baruch2021arkitscenes} to construct 5,000 QA pairs that span eight 3D spatial cognition tasks, categorized into three types of configurational, measurement estimation and spatiotemporal reasoning.
VLM4D~\cite{zhou2025vlm4d} uses both real-world clips and synthetic videos to produce over 2,000 high-quality QA pairs to explore models' spatiotemporal abilities in translational and rotational motion, perspective shifts, motion continuity, and related dynamics.
STI-Bench~\cite{li2025sti} is built from over 300 real-world videos with more than 2,000 QA pairs to test both static and dynamic spatial tasks.
More recently, VSI-SUPER~\cite{yang2025cambrian} complements VIS-Bench by adding long videos in the benchmark to construct a large instruction-tuning dataset named VSI-590K.
While these benchmarks emphasize various spatial capabilities in different scenarios, there is a lack of a comprehensive benchmark that unifies multi-task spatial understanding and reasoning in the context of a unified 4D representation.
Unlike existing benchmarks, Spatial4D-Bench is a large-scale, multi-task evaluation benchmark comprising \textasciitilde40,000 question-answer pairs covering 18 well-defined tasks which are systematically organized into 6 categories.
Spatial4D-Bench also presents seven tasks that are important for spatial intelligence but have not been sufficiently investigated in current benchmarks.
The statistical differences between Spatial4D-Bench and existing benchmarks are summarized in \cref{tab:recent_mllm_eval_benchmarks}.

\subsection{Multimodal Large Language Models}
Multimodal Large Language Models extend language models~\cite{gemini2023,llama2_2023} by integrating extra input modalities such as video, image, and audio, to enable advanced spatial capabilities including visual reasoning and scene understanding with temporal information~\cite{Alayrac2022Flamingo,chen2024internvl}.
Qwen-VL~\cite{bai2025qwen2,qwen3} focuses on architectural design to improve high-resolution visual recognition and OCR/text understanding inside images.
InternVL~\cite{wang2025internvl3,chen2024internvl} emphasizes scalable vision encoders and multimodal preference optimization to enable scalable training pipelines.
LLaVA~\cite{liu2023improvedllava} introduces vision instruction tuning to teach the model to follow complex multimodal instructions and makes it possible to train high-quality vision language models with relatively little manually curated multimodal data.
VideoLLama~\cite{zhang2025videollama} balances input flexibility and model efficiency to support long-video QA, temporal reasoning, temporal grounding.
These improvements jointly impel MLLMs to be more capable of achieving a wide variety of challenging spatial tasks.

\subsection{Spatial Reasoning with MLLMs}
In addition to benchmarking MLLMs, some researchers have also proposed various techniques to enhance the spatial reasoning capabilities of MLLMs.
SpatialRGPT~\cite{cheng2024spatialrgpt} develops automated data labeling pipelines to generate large quantities of 3D visual QA pairs for training, with the integration of depth modules and the alignment of depth information with 2D visual embeddings.
Tang \etal~\cite{tang2025sparkle} investigates whether enhancing basic spatial abilities such as direction comprehension, distance estimation, and localization can improve the overall spatial reasoning performance.
Chen \etal~\cite{chen2025spatial} reveals that successful spatial reasoning largely depends on the model's ability to attend to task-relevant objects.
Some other work~\cite{fan2025vlm,wu2025spatial,zhu2024llava} further lifts 2D encodings to 3D to recover implicit 3D structural information.
While various efforts have been made to improve the spatial reasoning capabilities of MLLMs, thorough evaluation of the model performance on a comprehensive benchmark will provide insights and understanding of how the current limitations can be further addressed.

\section{Spatial4D-Bench}
\label{sibench}

\subsection{Overview}
\label{sibench_overview}
Spatial4D-Bench is a large-scale, multi-task 4D spatial intelligence benchmark designed to assess the 4D spatial intelligence of MLLMs.
We name this benchmark ``Spatial4D-Bench'' as it is constructed from a large amount of video data grounded in 4D space and aims to evaluate the gap between MLLMs and human-level 4D spatial intelligence.
Spatial4D-Bench consists of \textasciitilde40,000 carefully curated and annotated QA pairs of 18 tasks, ranging from fundamental perception to complex reasoning. A detailed categorization of the 18 tasks is shown in \cref{fig_spatial4dbench_overview}.%

To ensure a comprehensive evaluation, Spatial4D-Bench aggregates data from a wide variety of publicly available datasets, encompassing both indoor and outdoor environments.
These datasets capture a rich variety of objects, actions, and scenes viewed from both egocentric and allocentric perspectives.
Instead of merely aggregating existing labels, we generate novel QA pairs across multiple datasets to maximize task diversity. 
Crucially, adhering to human spatial cognition principles~\cite{waller2013handbook, newcombe2004spatial}, we organize these 18 tasks into a hierarchical taxonomy composed of 6 core categories: \textit{object understanding}, \textit{scene understanding}, \textit{spatial relationship understanding}, \textit{spatiotemporal relationship understanding}, \textit{spatial
reasoning} and \textit{spatiotemporal reasoning}.
Compared to existing benchmarks, Spatial4D-Bench provides a more comprehensive evaluation suite to assess the spatial cognitive abilities of MLLMs, spanning a variety of tasks that parallel the versatility of human spatial intelligence.

\subsection{Benchmark Construction}
\label{subsec:construction}

\begin{figure} 
	\centering	
		\includegraphics[width=0.99\linewidth]{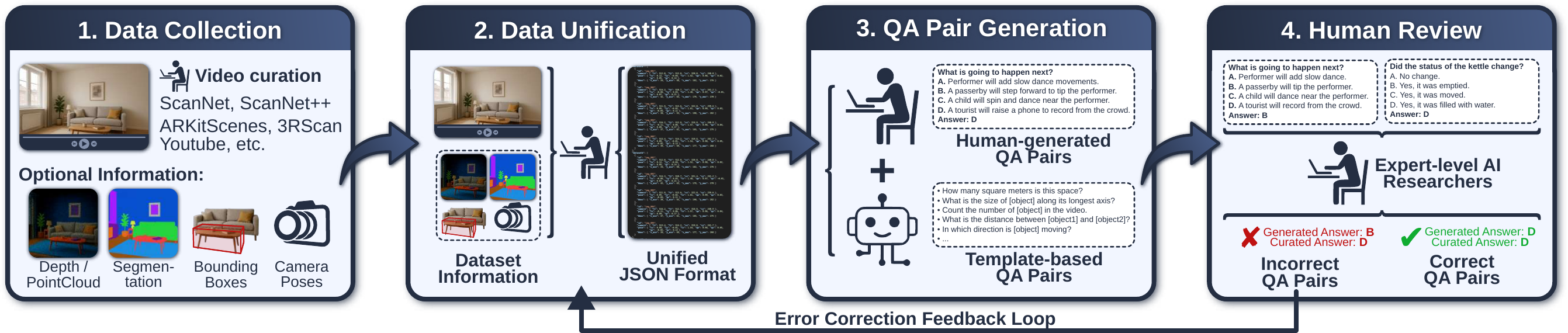}
	\caption{The Spatial4D-Bench construction pipeline. We unify diverse data sources into a standardized metadata format, generating QA pairs via templates and expert annotation, followed by rigorous human verification.   
	\label{benchmark_construction_diagr_v1} }
\end{figure}

\cref{benchmark_construction_diagr_v1} illustrates an overview of the pipeline used to construct Spatial4D-Bench.
It can be seen that there are four stages in the construction pipeline, including data collection, data unification, QA pairs generation, and final human review.

\begin{figure}
    \centering
    \includegraphics[width=0.99\linewidth]{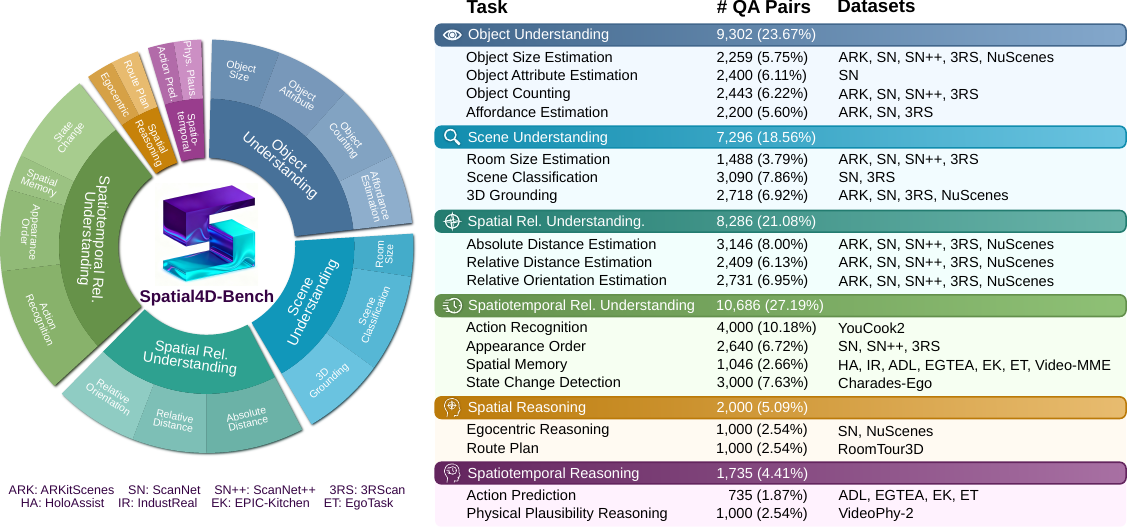}
    \caption{Distribution of question-answer pairs provided by our Spatial4D-Bench.}
    \label{fig:dataset_statistics}
\end{figure}

\paragraph{Data Collection.}
In the first stage, we collect a large amount of data from a variety of datasets, including Charades-Ego~\cite{sigurdsson2018charades}, ARKitScenes~\cite{baruch2021arkitscenes}, ScanNet~\cite{dai2017scannet}, ScanNet++~\cite{yeshwanth2023scannet++}, 3RScan~\cite{wald2019rio}, RoomTour3d~\cite{han2025roomtour3d}, EPIC-KITCHENS~\cite{Damen2018EPICKITCHENS}, ADL~\cite{youhome2022}, EgoTaskQA~\cite{jia2022egotaskqa}, HoloAssist~\cite{wang2023holoassist}, IndustReal~\cite{schoonbeek2024industreal}, YouCook2~\cite{zhou2018towards}, Video-MME~\cite{fu2024video},
VideoPhy-2~\cite{bansal2025videophy}, and nuScenes~\cite{caesar2020nuscenes}.
These datasets encompass various objects, actions, and scenes across indoor and outdoor scenarios, egocentric and allocentric perspectives, and both real and synthetic/generated data, contributing to the diversity and comprehensiveness of Spatial4D-Bench.
We aggregate diverse data modalities from the participating datasets, including text, RGB videos, and point clouds (used only to facilitate QA annotation).
\cref{fig:dataset_statistics} summarizes the source datasets that contribute to each task in Spatial4D-Bench.
In addition, as shown in~\cref{benchmark_construction_diagr_v1}, this stage includes a human verification loop to filter out low-quality scans or incomplete annotations, ensuring that the input data meet the reliability requirements.

\paragraph{Data Unification.}
In the second stage, we convert the collected data into a unified metadata structure following~\cite{Yang_2025_CVPR} since these data are collected from a variety of datasets and are heterogeneous.
This standardization facilitates consistent downstream task generation and enables traceability for error correction during human review and verification.
During this stage, we also preprocess image frames, sourced from RoomTour3D~\cite{han2025roomtour3d}, for the \textit{route plan} task by tagging room names on the relevant frames.
This preprocessing disambiguates the route planning targets, ensuring that the model focuses on spatial planning capabilities rather than OCR capabilities. For \textit{action prediction} and \textit{spatial memory} tasks, we select and extract video clips with various lengths from EPIC-KITCHENS~\cite{Damen2018EPICKITCHENS}, ADL~\cite{youhome2022}, EgoTaskQA~\cite{jia2022egotaskqa}, HoloAssist~\cite{wang2023holoassist}, IndustReal~\cite{schoonbeek2024industreal}, and Video-MME~\cite{fu2024video}, to satisfy the particular design requiremets of the QA pairs, achieving consistency between a QA pair and its corresponding video clip in the context of the given tasks. As a result, a subset of videos from the above-mentioned datasets have been preprocessed in curating Spatial4D-Bench.
Human review and verification are also used in this stage to ensure the correctness of data unification.

\paragraph{QA Pairs Generation.}
With the standardized metadata, we generate Question-Answer (QA) pairs via a combination of human annotation and automatic template-based generation.
We recruit several well-educated annotators with relevant backgrounds to design QA pairs for tasks including \textit{object attribute estimation, spatial memory, state change detection, egocentric reasoning, route plan, action prediction, and physical plausibility reasoning}.
Most of these tasks involve complex spatiotemporal understanding and reasoning in high-level cognition, making automatic template-based generation prone to producing low-quality QA pairs, whereas human annotation produces significantly higher-quality ones.
For the other tasks, we use automatic template-based generation based on well-designed template and ground-truth labels.
Specifically, for tasks involving rigid geometric properties, such as object counts, dimensions, and distances, we utilize template-based generation derived from the unified metadata.
And for tasks involving 3D boxes and coordinate systems, we provide multiple instruction formats such as object 3D bounding boxes, text descriptions of objects/rooms, or a combination of both, allowing foundation models to be tested with diverse instruction types.
Note that templates are also provided for annotating egocentric reasoning to assist human annotators.
Throughout this stage, humans continuously review and verify the generated QA pairs to judge, refine, correct, and filter out incorrect and ambiguous questions, options, and answers.

\paragraph{Final Human Review.}
In the final stage, experienced AI researchers perform the final human review to filter out ambiguous or incorrect QA pairs, \eg, checking whether a textual description uniquely identifies a target object.
As illustrated in \cref{benchmark_construction_diagr_v1}, errors detected at this stage will trigger a feedback loop where issues are traced back to the source metadata for correction.
Finally, we select \textasciitilde40,000 high-quality QA pairs to construct the Spatial4D-Bench.
Some examples of QA pairs can be found in the appendix (\cref{tab:example_qas_table_1} and \cref{tab:example_qas_table_2}).

\subsection{Task Taxonomy}
\label{subsec:taxonomy}

\begin{figure} 
	\centering	
		\includegraphics[width=0.99\linewidth]{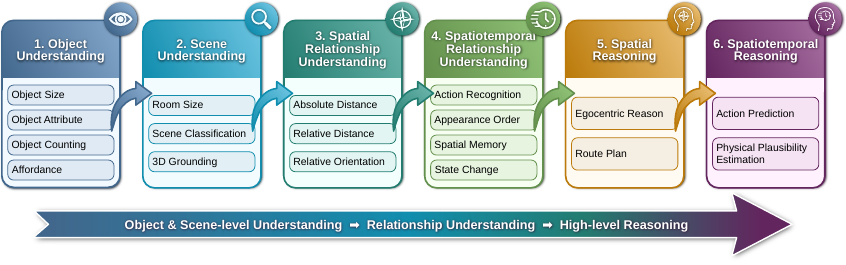} 
    \caption{Spatial4D-Bench Task Taxonomy.
    We organize 18 distinct tasks into 6 progressive categories representing the spectrum of spatial cognition.
    The taxonomy progresses from perception and understanding in object/scene level, through spatial/spatiotemporal understanding, to dynamic spatial/spatiotemporal reasoning, mirroring the cognitive abilities of human intelligence.}
     \label{fig:spatial_cogn_diagram}
\end{figure}

To systematically evaluate the gap between current MLLMs and human-level spatial cognition, as shown in \cref{fig:spatial_cogn_diagram}, we organize the 18 tasks into a hierarchical taxonomy composed of six core categories: \textit{object understanding}, \textit{scene understanding}, \textit{spatial relationship understanding}, \textit{spatiotemporal relationship understanding}, \textit{spatial reasoning} and \textit{spatiotemporal reasoning}. This taxonomy progresses from object/scene-level perception, through spatial/spatiotemporal understanding, to dynamic spatial/spatiotemporal reasoning, mirroring the cognitive abilities of human intelligence~\cite{waller2013handbook, newcombe2004spatial}.

\paragraph{Object Understanding.} Spatial intelligence begins with the accurate perception of intrinsic physical properties, which is a prerequisite for interacting with a physical scene.
This category includes \textit{object size estimation}, \textit{object attribute estimation},  \textit{object counting} and \textit{affordance estimation}. 
A robust spatial agent must ground visual features into both precise metric values (size) and semantic properties (attributes).
While existing benchmarks such as VSI-Bench~\cite{Yang_2025_CVPR} evaluate metric properties, they often overlook intrinsic attributes such as color, shape, and material.
Spatial4D-Bench integrates both metric and attribute understanding into a unified evaluation framework.
Crucially, we also incorporate \textit{affordance estimation} here as a functional attribute. 
Unlike SIBench~\cite{yu2025far} or SpaCE-10~\cite{gong2025space}, which focus on simple existence queries (\eg, ``is there an item to sit on?''), we integrate affordance with 3D grounding, requiring the model to identify the specific spatial position that satisfies a functional description (\eg, ``which position allows me to wash hands?''). 
This verifies that MLLMs possess a comprehensive understanding of an object's physical characteristics, which is a fundamental prerequisite for complex interaction tasks.

\paragraph{Scene Understanding.} Beyond isolated entities, an agent must comprehend the global semantic and geometric context of the environment to answer ``what'' is present and ``where'' it belongs.
This category encompasses \textit{room size estimation}, \textit{scene classification}, and \textit{3D grounding}.
Here, the model must abstract local visual cues into a broader understanding of the environment type and its volumetric scale.
Crucially, we include \textit{3D grounding} (object detection and localization in 3D space) in this category as it represents the population of the scene layout.
Unlike benchmarks like InternSpatial~\cite{deng2025internspatial} that rely on 2D bounding boxes, our \textit{3D grounding} task requires 3D spatial grounding, ensuring that MLLMs possess a volumetric understanding of object placement.

\paragraph{Spatial Relationship Understanding.} Once entities and the scene are defined, a spatial intelligence model needs to be able to resolve the geometric topology and metric layout between them.
This category is composed of \textit{relative distance estimation}, \textit{absolute distance estimation} and \textit{relative orientation estimation}.
Navigating the physical world requires resolving spatial ambiguities. By evaluating both relative (topological) and absolute (metric) distances, we test the robustness of the model's depth perception.
While benchmarks like SpatialRGPT~\cite{cheng2024spatialrgpt} address relative spatial relations, we enforce a more strict evaluation by combining relative topology with absolute metric estimation, ensuring that the model is not merely guessing based on 2D perspective cues.

\paragraph{Spatiotemporal Relationship Understanding.} Incorporating the fourth  dimension of time, this category evaluates the ability to track object states and dynamics over time. This category represents the leap from 3D to \textbf{4D spatial intelligence}.
Tasks include \textit{action recognition}, \textit{appearance order}, \textit{spatial memory}, and \textit{state change detection}.
The real world is a dynamic environment in which spatial and temporal aspects continuously evolve, and humans living in such an environment inherently develop 4D spatial intelligence that supports a wide range of spatial understanding and reasoning abilities.
While VSI-Bench includes basic appearance ordering~\cite{Yang_2025_CVPR} where clearly visible objects need to be identified in terms of the sequence in which they show up in a video, true 4D intelligence requires maintaining a coherent world model of objects even when they are occluded.
Our \textit{spatial memory} task evaluates this working memory capability by requiring the models to track objects that exit the field of view.
\textit{Action recognition} evaluates the model's ability to semantically categorize dynamic events within 3D space, bridging the gap between object detection and dynamic scene understanding.
Furthermore, \textit{state change detection} tests the causal understanding of interactions over time, such as a door opening rather than just movement tracking, distinguishing between simple movement and meaningful state transitions (e.g., a door opening).

\paragraph{Spatial Reasoning.} Moving from passive perception to active embodiment, this category assesses the agent's ability to reason about its own state and plan movements.
It is composed of \textit{egocentric reasoning} and \textit{route plan}.
We introduce \textit{egocentric reasoning} to challenge the model to infer the observer's own spatial state (\eg, ``how did the camera rotate?''), which is a prerequisite for self-localization.
This capability is critical for embodied agents operating in first-person views to localize themselves within the environment, moving beyond purely allocentric reasoning.
Building on this, our \textit{route plan} task elevates the difficulty beyond the single-room queries of VSI-Bench~\cite{Yang_2025_CVPR}.
Instead of using single-room single-step queries found in previous benchmarks~\cite{Yang_2025_CVPR}, we require long-horizon trajectory planning across multiple rooms, simulating the sequential cognitive load of a mobile robot operating in a complex environment.
To answer correctly, the model must predict a valid \textit{sequence} of actions (\eg, selecting the correct directions for multiple blanks in an instruction set) rather than a single directional move.

\paragraph{Spatiotemporal Reasoning.}
The highest level of spatial cognition involves abstracting perceptual data into predictive models and physical laws.
Tasks include \textit{action prediction} and \textit{physical plausibility reasoning}.
True spatial intelligence implies a ``world model'' that allows for prediction and physical intuition, capabilities largely ignored by existing benchmarks listed in \cref{tab:recent_mllm_eval_benchmarks}.
\textit{Action prediction} evaluates this capability by demanding that the model forecasts future events based on partial visual history.
Unlike understanding tasks that rely on retrospective classification, prediction requires causal inference: the model must synthesize observed dynamics with logical deduction to anticipate the likely intent of agents or the physical trajectory of objects.
Complementing this, \textit{physical plausibility reasoning} (using AI-generated videos of physical anomalies) tests whether the model has internalized the laws of physics, allowing it to identify violations of physical laws (\eg, gravity defiance), ensuring that their reasoning is grounded in reality.
This category benchmarks the transition from passive perception to active reasoning.

\section{Experiments}
\label{experiments}

\subsection{Evaluation Setup}
\label{sec:evaluation_setup}

\paragraph{Benchmark Models.}
We benchmark a diverse set of state-of-the-art (SOTA) multimodal large language models, comprising both leading proprietary models (GPT-5~\cite{OpenAI2025} and Gemini 2.5-Pro~\cite{Gemini2025}) and top-tier open-source models (VideoLLama3~\cite{zhang2025videollama}, Qwen2.5-VL~\cite{bai2025qwen2}, Qwen3-VL~\cite{bai2025qwen2}, and InternVL3.5~\cite{wang2025internvl3}).
To analyze the impact of model capacity, we evaluate variants across model sizes ranging from 7B to 241B parameters.
In total, we comprehensively evaluate eleven models on Spatial4D-Bench.
At the time of evaluation, these models represented the state-of-the-art.
We note that the results reported for proprietary models reflect the versions accessible during our experimental window; as these API-based services are subject to continuous updates by their providers, exact reproducibility may vary over time.
For memory-intensive models (\eg, 241B parameters), we adopt a uniform sampling strategy of $64$ frames from each video to mitigate memory constraints during evaluation following~\cite{Yang_2025_CVPR}.

\paragraph{Evaluation Protocol.}
We conduct all evaluations under a zero-shot setting.
Adhering to the protocol established in~\cite{Yang_2025_CVPR}, we compute the metrics tailored to the distinct answer formats in our Spatial4D-Bench, including Multiple-Choice Answers (MCA) and Numerical Answers (NA).
For MCA tasks, we report  exact matching with possible fuzzy matching.
For NA tasks, we evaluate performance using the Mean Relative Accuracy (MRA).
The MRA measures the model's consistency across a spectrum of error tolerances, calculated as the average satisfaction rate over a set of thresholds $\mathcal{C}$, which is defined as follows:
\begin{equation}
    \mathrm{MRA} = \frac{1}{|\mathcal{C}|}\sum_{\theta \in \mathcal{C} }\mathbbm{1} \left( \frac{\left | \hat{y} - y \right | }{y} < 1 - \theta \right), 
\end{equation}
where $\hat{y}$ and $y$ represent the predicted value and ground truth, respectively.
The threshold set is defined as $\mathcal{C} = \{0.5, 0.55, ... ,0.95\}$, representing a range of strictness levels.

\paragraph{Reference Baselines.}
Similar to~\cite{Yang_2025_CVPR}, to contextualize MLLM performance, we compare against the baselines of: \textit{Human Level Performance} and \textit{Chance Level Baselines (Random and Frequency)}. 
For \textit{Human Level Performance}, we sample 1,000 QA pairs from Spatial4D-Bench as a representative subset and recruited several qualified human evaluators with relevant research background to independently complete the test.
To establish a theoretical ceiling for the performance of human-level 4D spatial intelligence, we report the highest score among human evaluators. 
For \textit{Chance Level (Random)}, we report the expected average accuracy of random selection (MCA only).
For \textit{Chance Level (Frequency)}, we report performance by always selecting the dataset's most frequent answer for each task. This indicates the potential performance gains that could be obtained due to the long-tail distribution of answers or the imbalance in multiple-choice distributions.

\subsection{Main Evaluation Results}
\label{sec:main_results}
\begin{figure} 
	\centering
    \begin{subfigure}{0.37\textwidth}
        \centering
        \includegraphics[width=\linewidth]{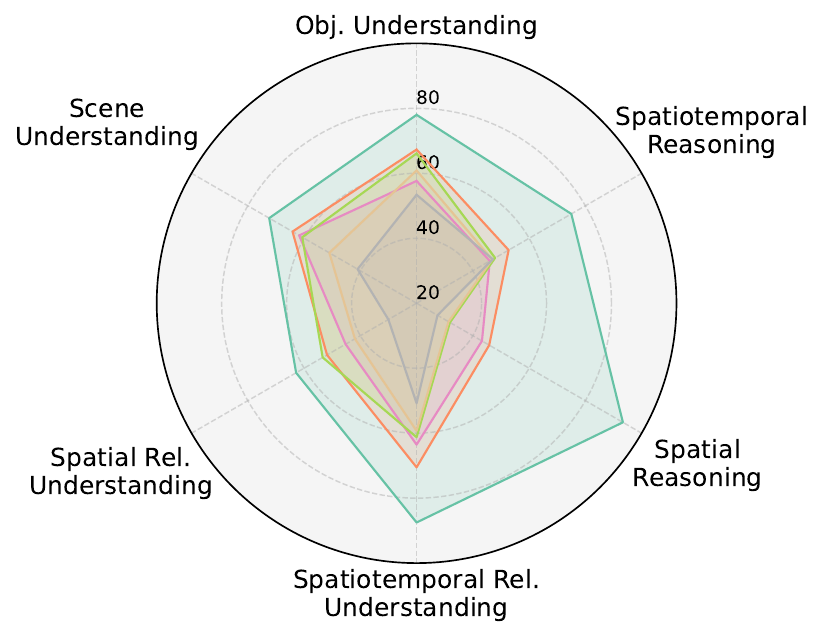}
        \caption{All categories}
        \label{fig:left}
    \end{subfigure}\hfill
    \begin{subfigure}{0.61\textwidth}
        \centering
        \begin{subfigure}[t]{0.32\textwidth}
            \centering
            \includegraphics[width=\linewidth]{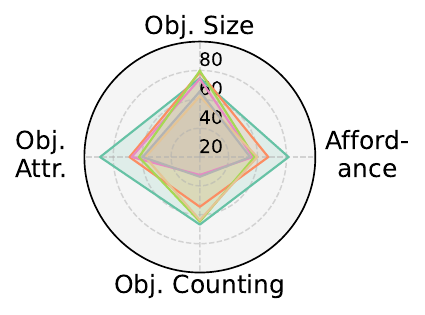}
            \vspace{-1.75em}
            \captionsetup{font=scriptsize}
            \caption{Obj. Understanding}
            \label{fig:1b}
        \end{subfigure}\hfill
        \begin{subfigure}[t]{0.32\textwidth}
            \centering
            \includegraphics[width=\linewidth]{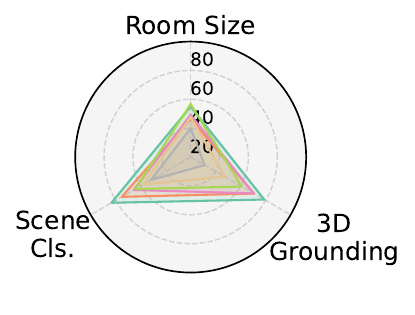}
            \vspace{-1.75em}
            \captionsetup{font=scriptsize}
            \caption{Scene Understanding}
            \label{fig:1c}
        \end{subfigure}\hfill
        \begin{subfigure}[t]{0.32\textwidth}
            \centering
            \includegraphics[width=\linewidth]{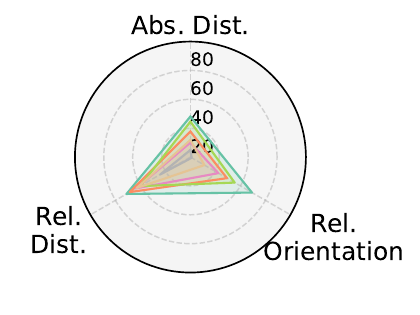}
            \vspace{-1.75em}
            \captionsetup{font=scriptsize}
            \caption{Spatial Rel. Understanding}
            \label{fig:2a}
        \end{subfigure}
        \begin{subfigure}[t]{0.32\textwidth}
            \centering
            \includegraphics[width=\linewidth]{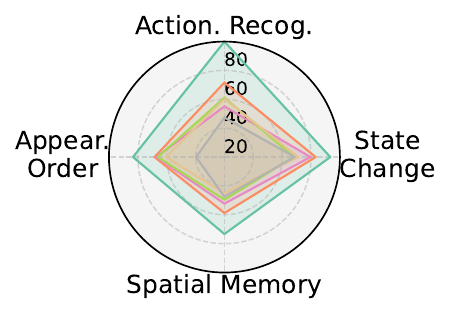}
            \vspace{-1.75em}
            \captionsetup{
                font=scriptsize,
                justification=centering,
                singlelinecheck=false
            }
            \caption{Spatiotemporal Rel.\\Understanding}
            \label{fig:2b}
        \end{subfigure}\hfill
        \begin{subfigure}[t]{0.32\textwidth}
            \centering
            \includegraphics[width=\linewidth]{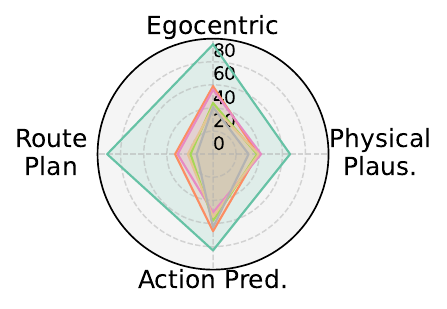}
            \vspace{-1.75em}
            \captionsetup{
                font=scriptsize,
                justification=centering,
                singlelinecheck=false
            }
            \caption{Spatial \& Spatio-\\temporal Reasoning}
            \label{fig:2c}
        \end{subfigure}\hfill
        \begin{subfigure}[t]{0.32\textwidth}
            \centering
            \vspace{-5em}
            \includegraphics[width=0.8\linewidth]{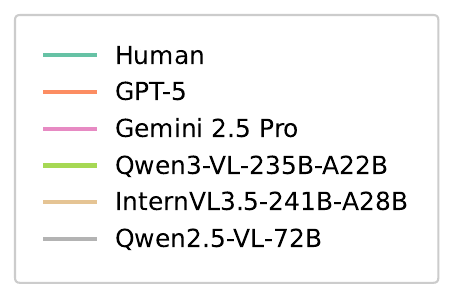}
        \end{subfigure}
        \vspace{-0.5em}
    \end{subfigure}
	\caption{Radar chart visualization of model performance across 18 tasks on Spatial4D-Bench.
    }
    \label{Main_results_radar_plot}
\end{figure}

We present the main evaluation results in \cref{table:main_eval_18_tasks} and show radar chart visualization of model performance across 18 tasks in \cref{Main_results_radar_plot}.
Our analysis and discussion of the main evaluation results are as follows.

\begin{table}
\centering
\scriptsize
\resizebox{\textwidth}{!}{
\begin{tabular}{llcccccccccccccc}
\toprule
& & & \multicolumn{2}{c}{\textit{Chance Level}} & \multicolumn{2}{c}{\textit{Proprietary Models}} & \multicolumn{9}{c}{\textit{Open-source Models}} \\
\cmidrule(lr){4-5} \cmidrule(lr){6-7} \cmidrule(lr){8-16}
 &  &  & &  & & & \multicolumn{2}{c}{\textbf{Qwen3-VL}} & \multicolumn{3}{c}{\textbf{InternVL3.5}} & \multicolumn{3}{c}{\textbf{Qwen2.5-VL}} & \textbf{VideoLlama3} \\
 \cmidrule(lr){8-9} \cmidrule(lr){10-12} \cmidrule(lr){13-15} \cmidrule(lr){16-16}
\multirow{-2}{*}{\textbf{Category}} & \multirow{-2}{*}{\textbf{Task}} & \multirow{-2}{*}{\textbf{Human}} & \multirow{-2}{*}{\textbf{Rand.}} & \multirow{-2}{*}{\textbf{Freq.}} & \multirow{-2}{*}{\textbf{GPT-5}}  & \multirow{-2}{*}{\textbf{\begin{tabular}[c]{@{}c@{}}Gemini 2.5\\ Pro\end{tabular}}} & \textbf{\tiny 235B-A22B} & \textbf{\tiny 30B-A3B} & \textbf{\tiny 241B-A28B} & \textbf{\tiny 38B} & \textbf{\tiny 8B} & \textbf{\tiny 72B} & \textbf{\tiny 32B} & \textbf{\tiny 7B} & \textbf{\tiny 7B} \\ \midrule
\multirow{4}{*}{Object Understanding} & Object Size & \cellcolor{tabgrey}{74.61} & - & - & 78.64 & 74.14 & 79.76 & \cellcolor{taborange}{80.10} & 62.17 & 65.63 & 47.39 & 65.47 & 62.27 & 35.63 & 33.21 \\
 & Object Attribute & 89.09 & 25.00 & 29.32 & \cellcolor{taborange}{68.71} & 67.25 & \cellcolor{tabgreen}{62.21} & 58.92 & 57.83 & 55.00 & 47.96 & 59.58 & 56.08 & 52.29 & 52.88 \\ 
& Object Counting & \cellcolor{tabgrey}{66.79} & - & - & 54.49 & 32.40 & 64.70 & \cellcolor{taborange}{67.74} & 63.63 & 60.59 & 55.99 & 33.65 & 37.89 & 46.84 & 52.69 \\ 
& Affordance & 81.48 & 25.00 & 28.27 & \cellcolor{taborange}{67.41} & 56.82 & 57.82 & 52.41 & 59.86 & \cellcolor{tabgreen}{63.36} & 49.41 & 54.86 & 51.00 & 40.00 & 33.91 \\ \midrule
\multirow{3}{*}{Scene Understanding} & Room Size & \cellcolor{tabgrey}{55.00} & - & - & 46.56 & 49.19 & 56.65 & \cellcolor{taborange}{67.22} & 47.62 & 55.02 & 50.16 & 39.97 & 50.35 & 39.82 & 28.41 \\
& Scene Classification & 83.33 & 25.00 & 26.11 & \cellcolor{taborange}{75.16} & 65.59 & \cellcolor{tabgreen}{64.38} & 54.72 & 58.06 & 61.18 & 48.40 & 51.81 & 52.32 & 42.71 & 50.86 \\
 & 3D Grounding & 78.85 & 25.00 & 33.90 & \cellcolor{taborange}{70.59} & 70.44 & \cellcolor{tabgreen}{60.88} & 50.55 & 47.35 & 38.09 & 33.86 & 31.14 & 29.41 & 25.96 & 26.99 \\ \midrule
\multirow{3}{*}{\begin{tabular}[l]{@{}l@{}}Spatial Relationship\\Understanding\end{tabular}} & Absolute Distance & 48.08 & - & - & 37.69 & 30.00 & \cellcolor{taborange}{44.52} & 42.20 & 31.80 & 28.51 & 25.89 & 24.90 & 28.15 & 18.93 & 23.84 \\
 & Relative Distance & 71.15 & 25.00 & 30.57 & \cellcolor{taborange}{68.57} & 63.37 & 60.23 & 58.11 & \cellcolor{tabgreen}{62.81} & 55.48 & 48.51 & 44.16 & 45.28 & 38.96 & 41.99 \\
 & Relative Orientation & 69.23 & 25.00 & 25.12 & 49.25 & 42.33 & \cellcolor{taborange}{55.40} & 53.94 & 31.16 & 52.47 & 40.17 & 21.02 & 42.29 & 24.79 & 31.01 \\ \midrule
\multirow{4}{*}{\begin{tabular}[l]{@{}l@{}}Spatiotemporal\\Relationship\\Understanding\end{tabular}} & Action Recognition & 100.00 & 25.00 & 25.80 & \cellcolor{taborange}{71.60} & 55.05 & \cellcolor{tabgreen}{61.12} & 42.95 & 60.32 & 48.55 & 40.37 & 47.33 & 43.53 & 39.25 & 42.90 \\
& Appearance Order & 83.33 & 25.00 & 26.14 & \cellcolor{taborange}{68.45} & 67.20 & \cellcolor{tabgreen}{66.17} & 61.48 & 60.53 & 54.77 & 50.27 & 39.70 & 37.20 & 39.36 & 40.27 \\
 & Spatial Memory &73.33 & 25.00 & 30.45 & \cellcolor{taborange}{58.80} & 52.29 & \cellcolor{tabgreen}{49.52} & 47.42 & 45.22 & 43.88 & 40.15 & 47.51 & 47.13 & 40.25 & 43.21 \\
 & State Change & 93.33 & 25.00 & 29.33 & \cellcolor{taborange}{83.20} & 79.57 & 68.20 & \cellcolor{tabgreen}{72.73} & 71.13 & 69.60 & 62.07 & 68.60 & 65.37 & 55.13 & 55.83 \\ \midrule
 \multirow{2}{*}{Spatial Reasoning} & Egocentric Reasoning & 95.00 & 25.00 & 32.57 & \cellcolor{taborange}{58.80} & 55.80 & \cellcolor{tabgreen}{44.20} & 41.90 & 40.90 & 36.80 & 31.10 & 40.70 & 34.90 & 36.00 & 40.80 \\
 & Route Plan & 91.67 & 5.03 & - & \cellcolor{taborange}{32.83} & 30.67 & 19.50 & 12.00 & \cellcolor{tabgreen}{21.83} & 15.50 & 9.83 & 14.17 & 14.17 & 13.50 & 14.67 \\ \midrule
\multirow{2}{*}{\begin{tabular}[l]{@{}l@{}}Spatiotemporal\\Reasoning\end{tabular}} & Action Prediction & 83.33 & 25.00 & 27.73 & \cellcolor{taborange}{66.67} & 50.48 & 57.69 & 56.46 & 54.29 & 45.47 & 42.99 & \cellcolor{tabgreen}{63.40} & 57.69 & 46.94 & 40.68 \\
 & Physical Plausibility & 66.67 & 25.00 & 30.10 & 38.78 & \cellcolor{taborange}{41.56} & 38.11 & 38.33 & \cellcolor{tabgreen}{39.44} & 36.33 & 29.22 & 30.78 & 29.89 & 31.89 & 35.22 \\ \midrule
 & \textit{Average} & 78.02 & - & - & \cellcolor{taborange}{60.90} & 54.68 & \cellcolor{tabgreen}{56.17} & 53.29 & 50.89 & 49.47 & 41.87 & 43.26 & 43.61 & 37.13 & 38.30 \\ \bottomrule
\end{tabular}
}
\setlength{\tabcolsep}{3pt}
\caption{Main evaluation results on Spatial4D-Bench.
\begin{tabular}[c]{c}\cellcolor{taborange}{Orange}\end{tabular} indicates best performance among all models, \begin{tabular}[c]{c}\cellcolor{tabgreen}{green}\end{tabular} indicates the best performance among open-source models, and \begin{tabular}[c]{c}\cellcolor{tabgrey}{grey}\end{tabular} indicates human performance that are surpassed by current MLLMs.}%
\label{table:main_eval_18_tasks}
\end{table}

\paragraph{The overall performance gap between MLLMs and human-level 4D spatial intelligence remains significant.}
From~\cref{table:main_eval_18_tasks} and \cref{fig:left}, we can see that the overall 4D spatial intelligence performance of MLLMs remains significantly inferior to that of humans.
The best proprietary model, GPT-5, achieves an average score of $60.90$, while the best open-source model, Qwen3-VL-235B-A22B, yields an average score of $56.17$.
This indicates that the performance gap between the best proprietary MLLMs and the best open-source MLLMs is relatively small, although the best proprietary MLLMs still perform better.
However, the overall performance gap between MLLMs (both proprietary and open-source) and human-level 4D spatial intelligence remains significant.
On Spatial4D-Bench, humans achieve an average score of $78.02$, outperforming the best proprietary model (GPT-5) by approximately $17$ and the best open-source model (Qwen3-VL-235B-A22B) by $22$.
Current MLLMs appear to operate as ``frame-based observers'' rather than ``world observers''.
They lack an intuitive physics engine or coherent temporal memory, preventing effective reasoning about causality, permanence, and dynamics.
Spatial4D-Bench effectively exposes this 4D reasoning gap, which was invisible in existing benchmarks.
Substantial efforts are required to further improve MLLMs toward human-level spatial cognition.

\paragraph{MLLMs have reached or even surpassed human-level spatial cognition on some understanding-related tasks.}
As shown in \cref{table:main_eval_18_tasks}, \cref{fig:1b} and \cref{fig:1c}, for the fundamental \textit{object/scene understanding} category (\eg~ \textit{object size}, \textit{object counting}, and \textit{room size}), existing MLLMs often reach or even surpass human performance.
In particular, Qwen3-VL-30B-A3B achieves $80.10$ on the object size estimation task, outperforming the human baseline of $74.61$.
In addition, on the \textit{object counting} task, the best MLLM also outperforms human performance. We attribute this to distinct cognitive processing differences. 
First, while humans rely on intuitive relative scale, they struggle with precise absolute metric estimation (e.g., ``is this table 1.2m or 1.4m?'') from 2D projections without explicit reference scales while MLLMs leverage massive prior knowledge obtained by pre-training on 3D and geometric data which are useful. Moreover, it is noticed that \textit{external factors} can also result in low accuracy of the human evaluation. For example, in \textit{object counting}, to elevate the the difficulty level in model evaluation, Spatial4D-Bench includes a set of low-resolution videos with significant jittering and inconsistent frames compared with videos used in~\cite{Yang_2025_CVPR}, posing a huge challenge to human evaluators in correctly matching and identifying the objects within these videos. We have also found that discrepancies among humans in defining the object categories also play an important role in the low accuracy of this task. While a \textit{nightstand} is considered to be a \textit{table} in the QA design, a human evaluator may not count it as a \textit{table}, causing inconsistent result with the ground truth answer.

\paragraph{MLLMs usually perform significantly worse than humans on spatial reasoning and spatiotemporal reasoning tasks.} The primary source of the human-AI gap lies in reasoning-related tasks.
As shown in \cref{table:main_eval_18_tasks}, \cref{fig:2b}, and \cref{fig:2c}, MLLMs exhibit substantial performance degradation on the \textit{spatial reasoning} and \textit{spatiotemporal reasoning} categories.
These challenging categories are the core of Spatial4D-Bench.
Specifically, for \textit{spatiotemporal reasoning}, even the top-tier MLLMs lag significantly behind humans (e.g., GPT-5 scores $32.83$ on \textit{route plan} vs.\ Human $91.67$).
This \textasciitilde$60\%$ gap highlights a fundamental deficiency of MLLMs in maintaining a coherent 4D world model over extended temporal sequences.
For the \textit{physical plausibility reasoning} task, MLLMs score near random chance ($30\%-40\%$), whereas humans intuitively reject these physically impossible scenarios ($66.67\%$).
This indicates that SOTA MLLMs struggle to ground visual perception in fundamental physical laws.

\begin{table}
\centering
\footnotesize
\begin{tabular}{c@{\extracolsep{4pt}}lcr@{\extracolsep{3pt}}lcr@{\extracolsep{3pt}}lcr@{\extracolsep{3pt}}l}
\toprule
 &  & \multicolumn{3}{c}{\textbf{Qwen3-VL {\tiny 30B-A3B}}} & \multicolumn{3}{c}{\textbf{InternVL3.5 {\tiny 38B}}} & \multicolumn{3}{c}{\textbf{InternVL3.5 {\tiny 8B}}} \\ 
 \cmidrule(lr){3-5} \cmidrule(lr){6-8} \cmidrule(lr){9-11}
 & \multirow{-2}{*}{\textbf{Task}} & \textbf{\scriptsize VSI} & \multicolumn{2}{c}{\textbf{\scriptsize Ours {\tiny ($\Delta$)}}} & \textbf{\scriptsize VSI} & \multicolumn{2}{c}{\textbf{\scriptsize Ours {\tiny ($\Delta$)}}} & \textbf{\scriptsize VSI} & \multicolumn{2}{c}{\textbf{\scriptsize Ours {\tiny ($\Delta$)}}} \\ \midrule
\multirow{4}{*}{\rotatebox[origin=c]{90}{\scriptsize \textit{Numerical}}} & Room Size & 65.17 & 67.22 & {\color{ForestGreen} (+2.05)} & 55.35 & 55.02 & {\color{BrickRed} (-0.33)} & 55.42 & 50.16 & {\color{BrickRed} (-5.26)} \\
 & Object Size & 77.35 & 80.10 & {\color{ForestGreen} (+2.75)} & 65.31 & 65.63 & {\color{ForestGreen} (+0.32)} & 61.19 & 47.39 & {\color{BrickRed} (-13.80)} \\
 & Object Counting & 71.50 & 67.74 & {\color{BrickRed} (-3.76)} & 65.38 & 60.59 & {\color{BrickRed} (-4.79)} & 61.31 & 55.99 & {\color{BrickRed} (-5.32)} \\
 & Absolute Distance & 42.46 & 42.20 & {\color{BrickRed} (-0.26)} & 31.02 & 28.51 & {\color{BrickRed} (-2.51)} & 34.40 & 25.89 & {\color{BrickRed} (-8.51)} \\ \midrule
\multirow{4}{*}{\rotatebox[origin=c]{90}{\scriptsize\textit{\begin{tabular}{@{}c@{}}Multiple\\-choice\end{tabular}}}} & Relative Distance & 54.93 & 58.11 & {\color{ForestGreen} (+3.18)} & 54.58 & 55.48 & {\color{ForestGreen} (+0.90)} & 50.42 & 48.51 & {\color{BrickRed} (-1.91)} \\
 & Relative Orientation & 58.78 & 53.94 & {\color{BrickRed} (-5.83)} & 60.02 & 52.47 & {\color{BrickRed} (-7.55)} & 44.21 & 40.17 & {\color{BrickRed} (-4.04)} \\
 & Route Plan & 41.75 & 12.00 & {\color{BrickRed} (-29.75)} & 37.63 & 15.50 & {\color{BrickRed} (-22.13)} & 34.02 & 9.83 & {\color{BrickRed} (-24.19)} \\
 & Appearance Order & 64.40 & 61.48 & {\color{BrickRed} (-2.92)} & 63.75 & 54.77 & {\color{BrickRed} (-8.98)} & 54.37 & 50.27 & {\color{BrickRed} (-4.10)} \\ \midrule
 & \textit{Average} & 59.54 & 55.23 & {\color{BrickRed} (-4.32)} & 54.13 & 48.50 & {\color{BrickRed} (-5.63)} & 49.42 & 41.03 & {\color{BrickRed} (-8.39)} \\ \bottomrule
\end{tabular}%

\caption{Comparison with VSI-Bench~\cite{Yang_2025_CVPR} on overlapping tasks.
Overall, state-of-the-art models perform worse on our Spatial4D-Bench, especially on the redefined \textit{route plan} task.
These results indicate that our Spatial4D-Bench is more challenging than VSI-Bench~\cite{Yang_2025_CVPR} on overlapping tasks.
}
\label{table:vsi_calib}
\end{table}

\subsection{Benchmark Challenge Analysis}
As summarized in~\cref{tab:recent_mllm_eval_benchmarks}, Spatial4D-Bench provides a more comprehensive evaluation benchmarks with 18 tasks.
Since there are some tasks overlapping with existing benchmarks, it would be interesting to investigate the challenge of these tasks between existing benchmarks and Spatial4D-Bench.
To this end, we compare with VSI-Bench~\cite{Yang_2025_CVPR}, one of the most representative spatial intelligence benchmarks, on overlapping tasks using Qwen3-VL-30B-A3B, InternVL3.5-38B, and InternVL3.5-8B.
As shown in~\cref{table:vsi_calib}, overall, state-of-the-art models perform worse on our Spatial4D-Bench.
Specifically, on some perception-related tasks, \eg, \textit{object size} and \textit{room size}, the performance is comparable across both benchmarks.
However, for more challenging reasoning tasks, such as \textit{relative orientation} and \textit{route plan}, substantial divergences can be observed, especially on the \textit{route plan} task, where the tested models exhibit a substantial performance drop of approximately 22\% to 29\% on Spatial4D-Bench compared to VSI-Bench~\cite{Yang_2025_CVPR}.
In addition to the difficulty elevation described in \cref{subsec:taxonomy}, this can be attributed to the fact that the \textit{route plan} task in Spatial4D-Bench requires long-horizon planning and 4D spatial understanding over longer sequences and more complex state transitions, which distinguishes Spatial4D-Bench from existing benchmarks that may rely on shorter horizons or simpler topological graphs.

\subsection{Further Analysis and Discussion}

\paragraph{Temporal Context and Spatial Memory.}
\textit{Spatial memory} in long video represents a frontier of 4D spatial intelligence, assessing the ability of a model to maintain consistent visual state representations over extended durations.
To study the impact of the sequence length, we evaluate Qwen3-VL-30B-A3B on videos of varying lengths ($5$, $10$, and $30$ minutes).
The evaluated subset comprises $420$, $426$, and $200$ QA pairs for the respective durations, totaling $1,046$ samples.
\begin{wraptable}{r}{0.4\textwidth}
    \centering
    \begin{tabular}{llll}	 
    \toprule 
         \textbf{Video duration}     & \textbf{Accuracy (\%)$\uparrow$} \\         
         \midrule
         5 min     & 0.5381 \\
         10 min    & 0.4343 \\
         30 min    & 0.4250 \\   
    \bottomrule    
	\end{tabular}
    \caption{Impact of video duration on \textit{spatial memory} accuracy (Qwen3-VL-30B-A3B). Performance degrades on longer sequences, highlighting the limitation of fixed-frame sampling in capturing high-frequency spatial updates.}
    \label{table:spatial_memory_video_length}
\end{wraptable}
We present the results in \cref{table:spatial_memory_video_length}.
As expected, performance drops as the video length increases, indicating that the information retrieval architecture can play an important role in the model's reasoning capability, as unified sampling scheme in longer videos will cause more information loss. However, the gap between 10-minute and 30-minute videos is much closer than that between 5-minute and 10 minute videos. We explain it as the bottleneck saturation from temporal aliasing in sampled frames, and the strong language prior will compensate for the information loss from visual input as will be shown later in \cref{table:visual_ablation}. Nonetheless, the result still indicates that solving 4D spatial intelligence requires a paradigm shift from fixed-context windows to adaptive sampling or streaming memory architectures.

\begin{table}[]
\centering
\footnotesize
\begin{tabular}{llcr@{\extracolsep{3pt}}lr@{\extracolsep{3pt}}lc}
\toprule
\textbf{Category} & \textbf{Task} & \textbf{Video Input} & \multicolumn{2}{c}{\textbf{Image Input {\tiny($\Delta$)}}} & \multicolumn{2}{c}{\textbf{Text Input {\tiny($\Delta$)}}}\\ \midrule
\multirow{4}{*}{Object Understanding} & Object Size & 80.10 & 59.92 & {\color{BrickRed}(-20.18)} & 48.48 & {\color{BrickRed}(-31.62)} \\
 & Object Attribute & 58.92 & 48.63 & {\color{BrickRed}(-10.29)} & \textbf{50.63} & {\color{BrickRed}(-8.29)} \\
 & Object Counting & 67.74 & 20.76 & {\color{BrickRed}(-46.98)} & 1.23 & {\color{BrickRed}(-66.51)} \\ 
 & Affordance & 52.41 & 42.91 & {\color{BrickRed}(-29.50)} & 26.14 & {\color{BrickRed}(-26.27)} \\ \midrule
\multirow{3}{*}{Scene Understanding} & Room Size & 67.22 & 12.28 & {\color{BrickRed}(-54.94)} & \textbf{27.47} & {\color{BrickRed}(-39.75)} \\
& Scene Classification & 54.72 & 36.93 & {\color{BrickRed}(-17.79)} & \textbf{44.62} & {\color{BrickRed}(-10.11)} \\
 & 3D Grounding & 50.55 & 41.54 & {\color{BrickRed}(-9.01)} & \textbf{45.40} & {\color{BrickRed}(-5.15)} \\ \midrule
\multirow{3}{*}{\begin{tabular}[c]{@{}l@{}}Spatial Relationship\\Understanding\end{tabular}} & Absolute Distance & 42.20 & 24.86 & {\color{BrickRed}(-17.34)} & 22.29 & {\color{BrickRed}(-19.91)} \\
 & Relative Distance & 58.11 & 38.50 & {\color{BrickRed}(-19.61)} & 34.01 & {\color{BrickRed}(-24.10)} \\
 & Relative Orientation & 53.94 & 39.11 & {\color{BrickRed}(-13.84)} & 10.80 & {\color{BrickRed}(-42.15)} \\ \midrule
\multirow{4}{*}{\begin{tabular}[c]{@{}l@{}}Spatiotemporal\\ Relationship\\Understanding\end{tabular}} & Action Recognition & 42.95 & 35.10 & {\color{BrickRed}(-7.85)} & 20.38 & {\color{BrickRed}(-22.57)} \\
 & Appearance Order & 61.48 & 29.96 & {\color{BrickRed}(-31.32)} & \textbf{31.36} & {\color{BrickRed}(-30.12)} \\
 & Spatial Memory & 47.42 & 35.66 & {\color{BrickRed}(-11.76)} &  \textbf{37.19} & {\color{BrickRed}(-10.23)} \\
 & State Change & 72.73 & 42.43 & {\color{BrickRed}(-30.30)} & 40.13 & {\color{BrickRed}(-32.60)} \\ \midrule
 \multirow{2}{*}{Spatial Reasoning} 
 & Egocentric Reasoning & 41.90 & 28.40 & {\color{BrickRed}(-13.50)} & 21.20 & {\color{BrickRed}(-20.70)} \\
 & Route Plan & 12.00 & 10.67 & {\color{BrickRed}(-1.33)} & \textbf{13.67} & {\color{ForestGreen}(+1.67)} \\ \midrule
\multirow{2}{*}{\begin{tabular}[l]{@{}l@{}}Spatiotemporal\\Reasoning\end{tabular}} 
 & Action Prediction & 56.46 & 37.14 & {\color{BrickRed}(-19.32)} & \textbf{39.86} & {\color{BrickRed}(-16.60)} \\
 & Physical Plausibility & 38.33 & 28.00 & {\color{BrickRed}(-10.33)} & 23.11 & {\color{BrickRed}(-15.22)} \\ \midrule
 & \textit{Average} & 53.29 & 34.04 & {\color{BrickRed}(-19.24)} & 29.89 & {\color{BrickRed}(-23.40)} \\ \bottomrule
\end{tabular}
\caption{Visual Ablation Study evaluated on Qwen3-VL-30B-A3B. \textbf{Video Input}: Full 64 frames. \textbf{Image Input}: Single frame (random). \textbf{Text Input}: Text-only. While performance generally degrades as visual information is removed, text-only input (bolded) outperforms single-frame input in tasks requiring global context (e.g., \textit{route plan}, \textit{scene classification}), suggesting that incomplete visual data can act as a distractor that overrides correct language priors.}

\label{table:visual_ablation}
\end{table}

\paragraph{Visual Ablation: Disentangling Perception from Language Priors.}
To assess the genuine spatial-temporal reasoning capabilities of MLLMs versus their reliance on language shortcuts, we conduct a visual ablation study on Qwen3-VL-30B-A3B.
We compare the standard \textbf{Video Input} ($64$ frames) against two baselines: \textbf{Single Frame} (a randomly selected image) and \textbf{Text Only} (blind evaluation with no visual input).
The results shown in Table~\ref{table:visual_ablation} reveal distinct performance patterns.

\begin{itemize}
    \item{
        \textbf{The Necessity of 4D Signals.}
        The significant performance gap between video and text-only inputs confirms the validity of Spatial4D-Bench.
        On average, removing visual signals causes a performance drop of $23.40\%$.
        This degradation is most pronounced in dynamic tasks such as \textit{state change} ($-32.60\%$) and \textit{object counting} ($-66.51\%$), demonstrating that these tasks require explicit temporal integration and cannot be solved via language models alone. For example, the model simply outputs $0$ for every question in \textit{object counting} without visual input.
    }
    \item{
        \textbf{The ``Blind'' Leading the ``One-Eyed''.}
        A distinct phenomenon emerges when comparing Text Only vs. Single Frame inputs.
        On eight specific tasks, \ie, \textit{object attribute, room size, scene classification, 3D grounding, appearance order, spatial memory, route plan} and \textit{action prediction}, the blind text-only baseline outperforms the single-frame input.
        For example, on \textit{scene classification}, text-only input achieves $44.62\%$ compared to $36.93\%$ for single-frame.
        We attribute this anomaly to two factors:
        \begin{enumerate}
            \item {\textbf{Language Priors vs.\ Random Chance:}
            In the absence of visual data, the model defaults to learned LLM priors, \eg, probabilistically associating ``oven'' with ``kitchen''.
            As similarly observed on VSI-Bench~\cite{Yang_2025_CVPR}, these priors often provide a statistical floor that is significantly better than random chance.}
            \item \textbf{The ``Misleading Frame'' Hypothesis:}
            For tasks requiring global or temporal continuity (\eg, \textit{room size} and \textit{route plan}), a single random frame often acts as an adversarial distractor.
            If the question concerns an entire apartment but the random frame depicts a small bathroom, the visual signal actively contradicts the correct answer, overriding the model's correct language prior.
        \end{enumerate}
        This inversion highlights a critical limitation in existing MLLMs: they struggle to weigh conflicting evidence, often letting incomplete visual data override reliable language priors.
    }
    \item{
        \textbf{The Route Plan Failure Case.}
        Most alarmingly, providing full video input for \textit{route plan} yields negligible improvement over the blind baseline (12.00\% vs 13.67\%).
        This suggests that current MLLMs struggle to construct a coherent spatial map from egocentric videos, effectively reverting to random guessing or language priors even when visual data are provided.
    }
\end{itemize}

\begin{figure}[t!]
	\centering	
	\includegraphics[width=\columnwidth]{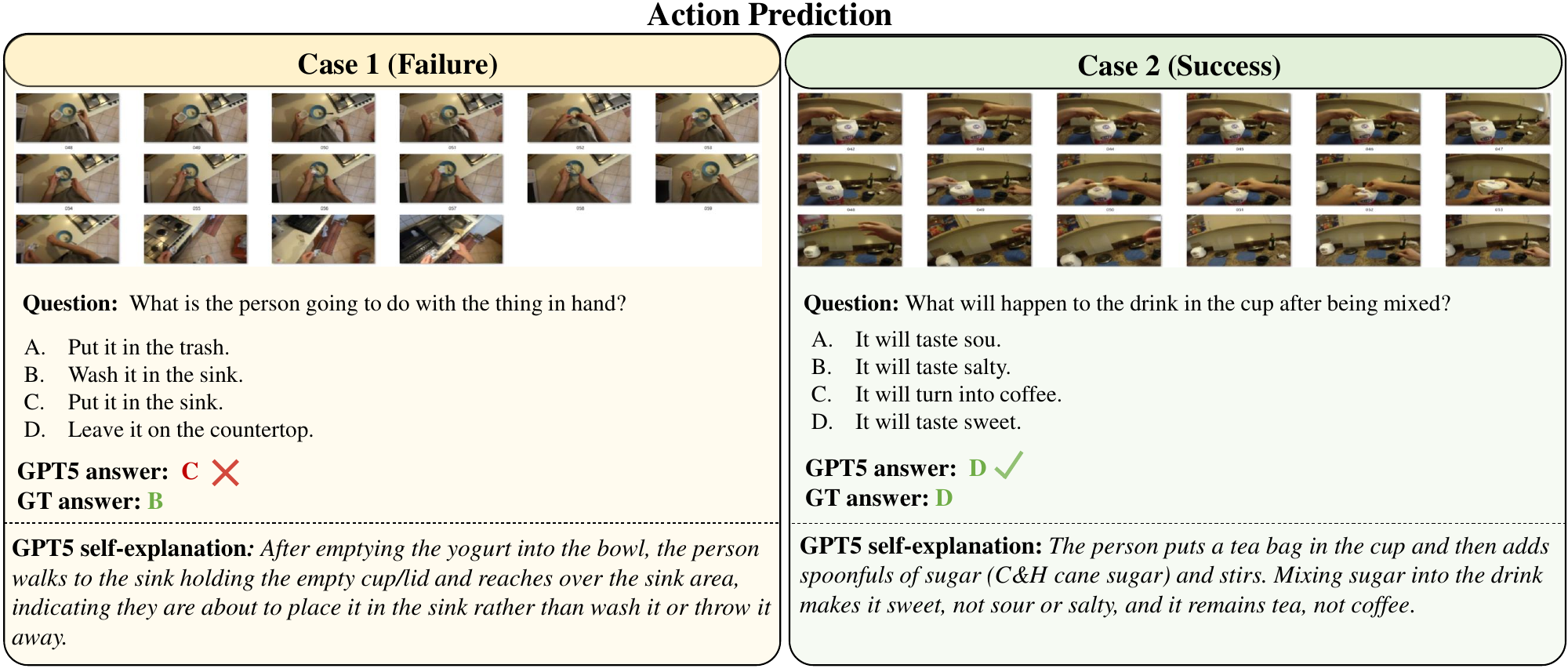} 		
	\caption{Action prediction examples (GPT-5). Spatial4D-Bench exposes the conflict between general semantic priors and specific visual evidence (Case 1), while validating successful text-driven reasoning (Case 2).
	\label{GPT5_spatial_time_causal_prediction_examples} }
\end{figure}

\subsection{Qualitative Case Analysis}
\label{sec:failure_analysis}

Our Spatial4D-Bench has exposed the critical bottlenecks of state-of-the-art MLLMs, especially on \textit{spatial reasoning} and \textit{spatiotemporal reasoning}.
However, quantitative metrics often mask the underlying reasoning processes. To uncover \textit{why} models fail, we analyze the failure and success cases of GPT-5, the top-performing MLLM on our evaluation, and leverage its self-explanation capabilities to trace the precise reasoning pathways that lead to errors.
In this way, we showcase how Spatial4D-Bench successfully exposes the systemic bottlenecks of MLLMs in maintaining temporal coherence, physical grounding, and perceptual fidelity that remain hidden in existing spatial intelligence benchmarks.

\paragraph{MLLMs are Relatively Fragile in Spatiotemporal Continuity.}
A major challenge in 4D spatial intelligence is \textit{spatiotemporal reasoning} that requires understanding a coherent 4D world over time rather than treating videos as a collection of disjoint semantic concepts.
By designing tasks that connect past, present, and future visual observations, Spatial4D-Bench reveals a significant \textit{temporal incoherence} in SOTA MLLMs.
\cref{GPT5_spatial_time_causal_prediction_examples} shows failure and success cases from the \textit{action prediction} task, which exposes this fragility.
In \textbf{Case 1} (failure), the ground truth action is ``washing'' a yogurt cup, but the model incorrectly predicts ``placing it in the sink'', driven by a semantic prior rather than the specific temporal action.
The benchmark's design allows us to probe the cause via the model's self-explanation: \textit{``...the person walks to the sink holding the empty cup... indicating they are about to place it in the sink rather than wash it or throw it away''}.
This textual feedback from the model demonstrates that the model successfully tracked the trajectory but failed to infer the latent intent (washing for recycling), proving that even powerful models fail to ground their predictions in the actual temporal dynamics when those dynamics conflict with training priors.
On the other hand, \textbf{Case 2} (success) validates the benchmark's ability to measure robust chain-of-thought reasoning when explicit cues (OCR) are present.
The model correctly infers the drink will taste ``sweet'' by synthesizing the label ``sugar'' with the stirring action (\textit{``mixing sugar into the drink makes it sweet...''}).
This contrast underscores the diagnostic value of our Spatial4D-Bench: it can distinguish between scenarios where models rely on robust textual grounding versus those where they collapse into hallucination due to temporal ambiguity.

\begin{figure}[t]
	\centering	
	\includegraphics[width=\columnwidth]{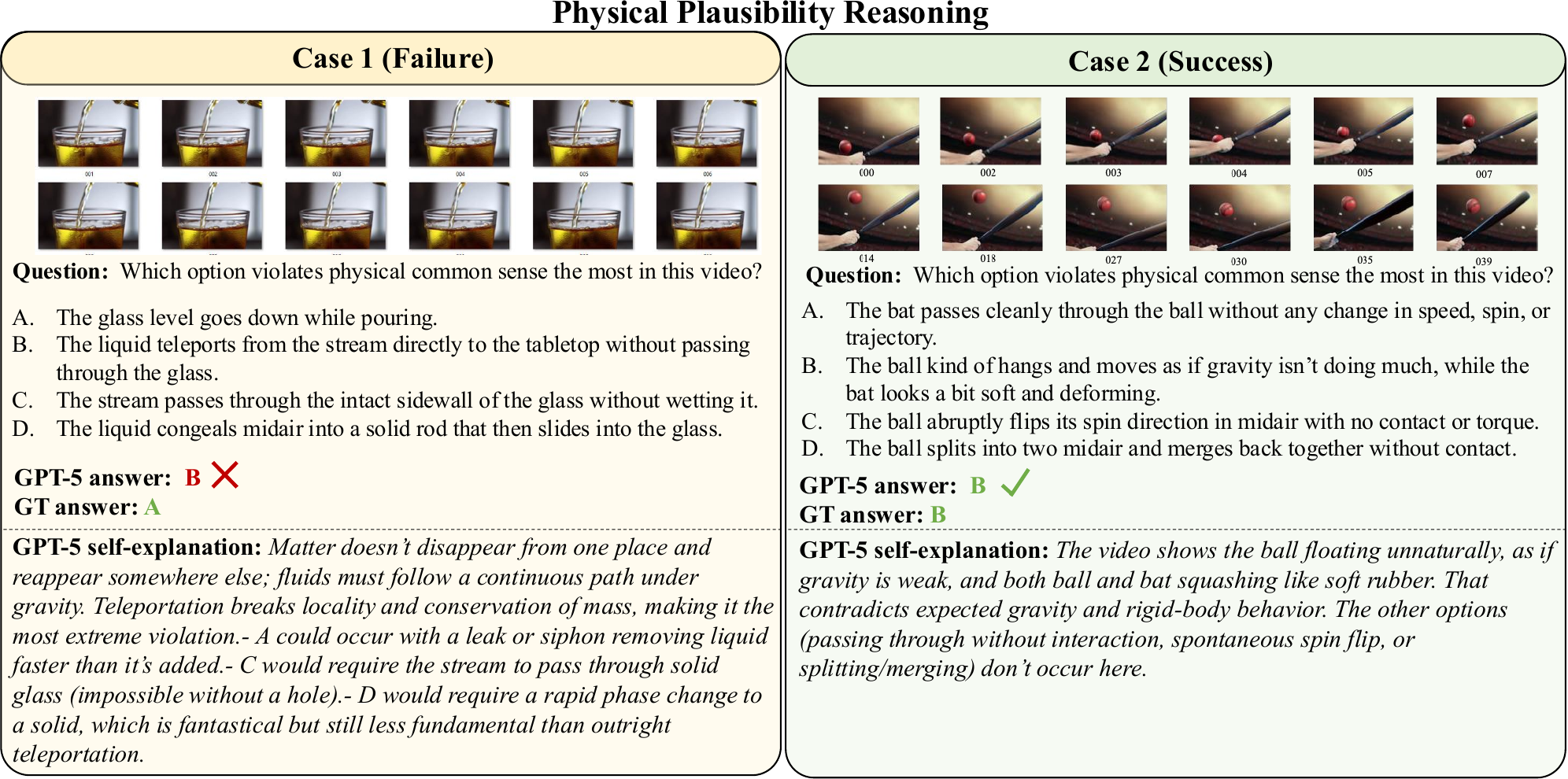}
	\caption{Physical plausibility reasoning examples (GPT-5). Spatial4D-Bench reveals that models can hallucinate physical justifications (Case 1) despite possessing correct theoretical knowledge, highlighting the gap between semantic and visual understanding.
	\label{GPT5_physical_common_sense_examples} }
\end{figure}

\paragraph{MLLMs Exhibit a Knowledge-Perception Gap in Intuitive Physics.}
By incorporating the \textit{physical plausibility reasoning} in the \textit{spatiotemporal reasoning} category, specifically using AI-generated physical anomalies, Spatial4D-Bench provides a unique evaluation of the ``intuitive physics engine'' of MLLMs.
Our evaluation exposes a sharp dissociation between low-level perceptual grounding and high-level physical knowledge.
In \cref{GPT5_physical_common_sense_examples} \textbf{Case 1} (failure), the benchmark challenges the model with a subtle violation of fluid dynamics.
GPT-5 fails to identify a physical anomaly where liquid levels behave inconsistently during pouring, yet its self-explanation recites perfect high-level physics principles: \textit{``Matter doesn’t disappear from one place and reappear somewhere else... Teleportation breaks locality and conservation of mass...''}.
This result highlights a critical insight enabled by our benchmark: SOTA models possess abstract knowledge of physical laws but lack the visual grounding to detect their violation in pixel space.
The model ``knows'' the laws of physics but cannot ``see'' them being broken, relying instead on analyzing the textual plausibility of the options (e.g., rejecting ``teleportation'' as a concept) rather than verifying the visual dynamics.
Spatial4D-Bench thus serves as a necessary filter to differentiate between models that merely \textit{know} physics textually and those that can \textit{perceive} physics visually.
In contrast, \textbf{Case 2} (success) shows that the model is capable of detecting violations in rigid body dynamics. It correctly identifies that a baseball and bat exhibit unnatural softness and defy gravity, suggesting that conspicuous deviations in material properties and trajectory dynamics are easier for current architectures to flag than subtle fluid inconsistencies.

\begin{figure}[t]
	\centering	
\includegraphics[width=\columnwidth]{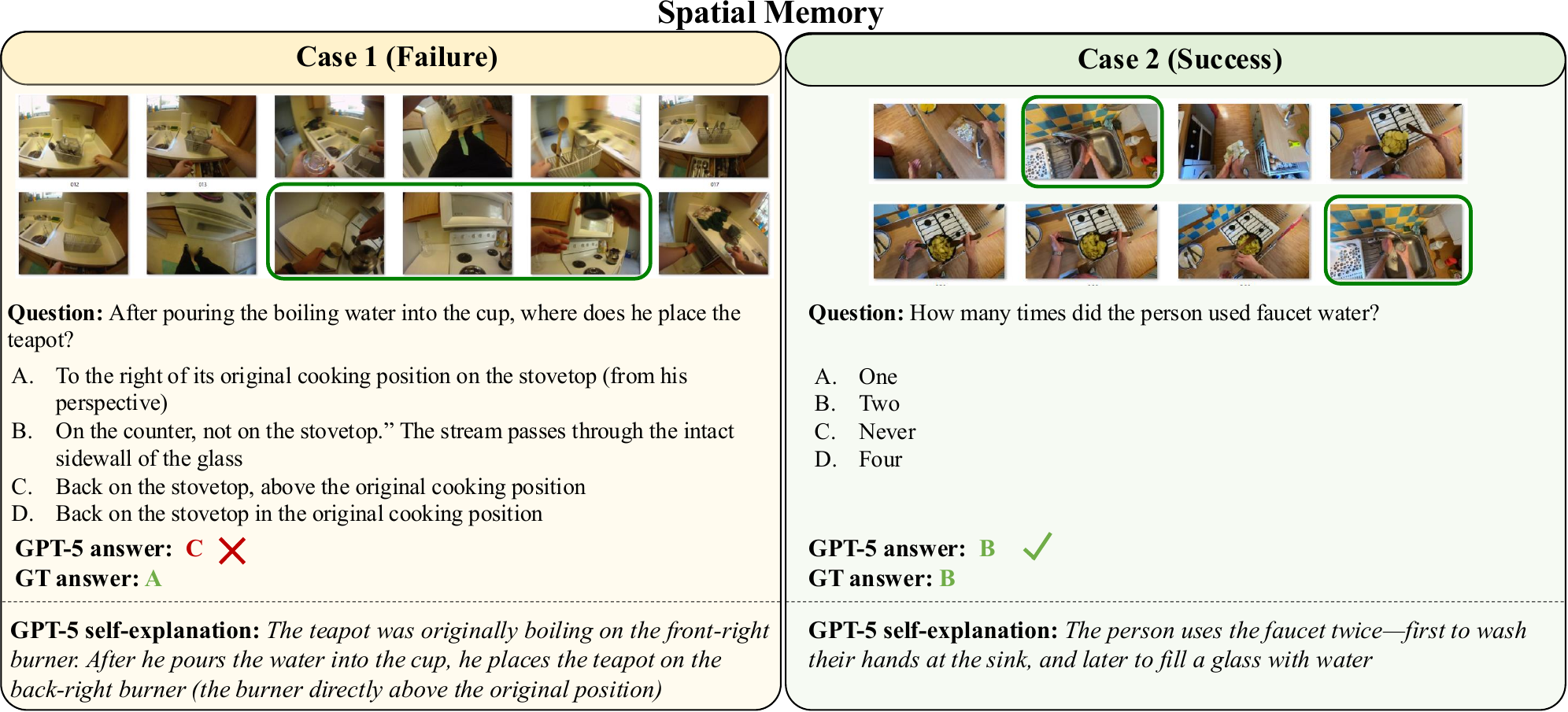} 		
	\caption{The benchmark's fine-grained tracking requirements expose confident hallucinations driven by texture confusion (Case 1).
	\label{GPT5_spatial_memory_failure_examples} }
\end{figure}

\begin{figure}[t]
	\centering	
	\includegraphics[width=\columnwidth]{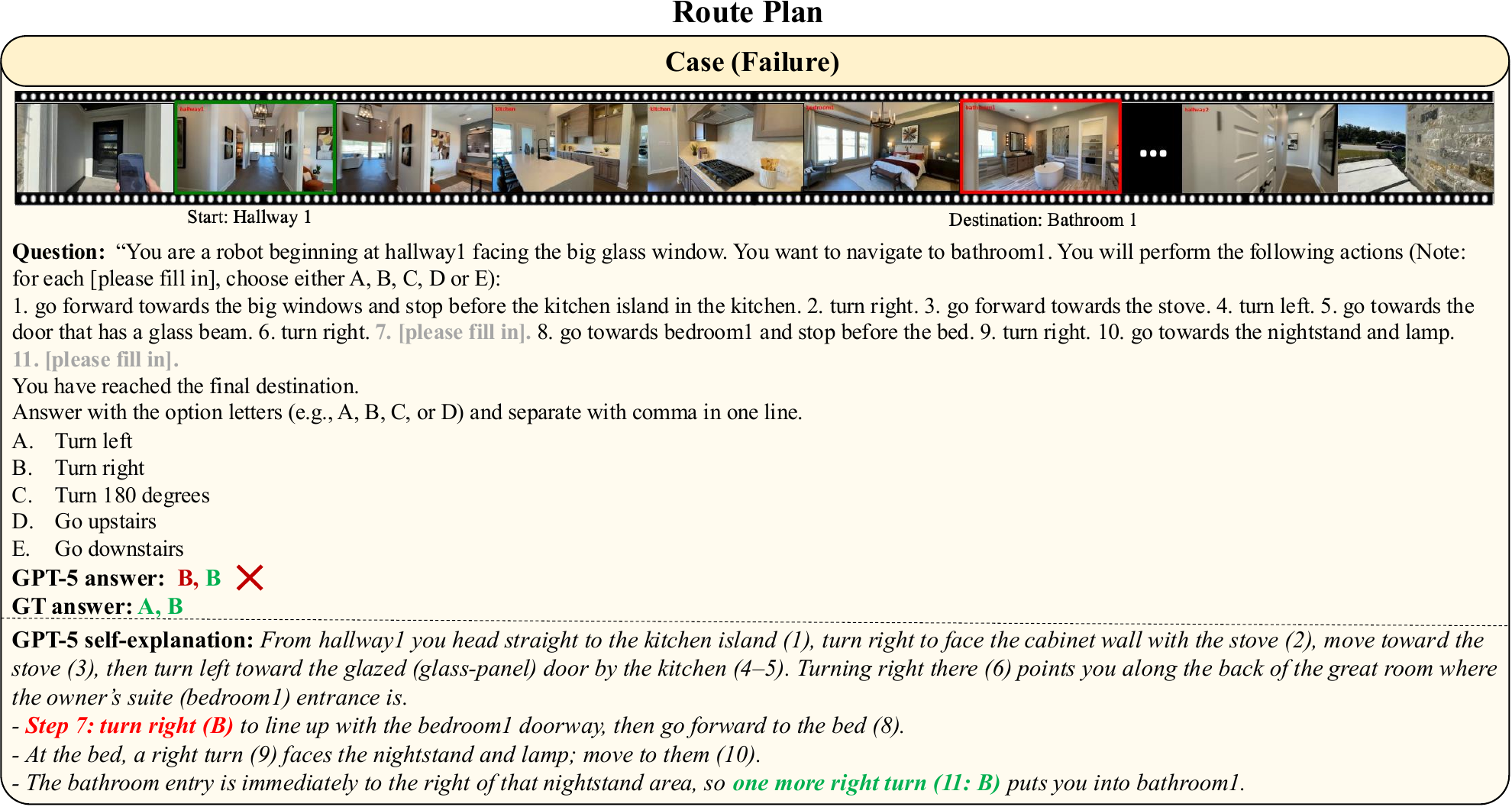}	
    \caption{Route plan failure case (GPT-5). 
    The model attempts to plan a path from the hallway to the bathroom but fails to ground its reasoning in the scene's geometry. 
    At step 7, the model confidently justifies a right turn (option B) to enter the bedroom, hallucinating a spatial layout that fits its internal narrative. 
    However, the visual evidence dictates a left turn (option A) is required to face the doorway. 
    This illustrates a critical limitation where SOTA models fail to maintain visual spatial consistency in long-horizon planning.}
	\label{fig:GPT5_navigation_failure} 
\end{figure}

\paragraph{MLLMs have Spatial Hallucination Driven by Texture Confusion.}
Through evaluation on Spatial4D-Bench, we observe that perceptual ambiguity can lead to confident hallucinations, even in tasks (such as \textit{spatial memory}) that require explicit tracking.
In \cref{GPT5_spatial_memory_failure_examples} \textbf{Case 1} (failure), the model incorrectly localizes the final position of a teapot. 
Crucially, the self-explanation reveals a confident hallucination: \textit{``After he pours the water... he places the teapot on the back-right burner (the burner directly above the original position)''}.
This contradicts the visual evidence of the teapot's placement in other locations.
This suggests that texture similarity across the stove top surface causes the model to lose track of the object's specific geometric coordinates.
Unlike a tracking failure where a model might express uncertainty, here the model constructs a coherent (but false) narrative to fill the perceptual gap.
This failure mode validates the necessity of Spatial4D-Bench's fine-grained annotation: unlike simpler existence or classification tasks, our spatial memory queries force the model to confront texture confusion and occlusion.
The benchmark demonstrates that even when a model is confident and generates plausible-sounding narratives, it often lacks the precise metric grounding required for 4D spatial intelligence.
By contrast, \textbf{Case 2} (success) demonstrates that the model maintains robust temporal tracking when visual events are semantically distinct.
The model correctly counts that the faucet was used ``two'' times.
Its self-explanation, \textit{``...first to wash their hands at the sink, and later to fill a glass with water''}, shows a successful linking of two separate temporal events, implying that the ``grounding gap'' is highly sensitive to visual saliency.
The model fails when tracking requires resolving low-level texture ambiguity (Case 1) but succeeds when tracking high-level, semantically distinct actions (Case 2), further validating the benchmark's ability to probe the granular limits of spatial memory.

\paragraph{MLLMs Rely on Hallucination in Egocentric Route Plan.}
The \textit{route plan} task in the \textit{spatial reasoning} category illustrates the inability of current MLLMs to construct accurate mental maps from egocentric video streams. 
In \cref{fig:GPT5_navigation_failure}, the model attempts to plan a path from a hallway to a bathroom. 
While GPT-5 outputs a confident self-explanation, reasoning that a right turn at step 7 would ``line up with the bedroom doorway'', this contradicts the visual geometry: the robot has just entered a hallway configuration where a left turn (option A) is geometrically required to face the bedroom entrance. 
The model hallucinates a spatial layout that fits its internal narrative but ignores the visual reality of the scene. 
Furthermore, while the model correctly identifies the final turn (step 11) as a right turn, this success is coincidental, derived from a flawed intermediate trajectory. 
This failure highlights a critical bottleneck: SOTA MLLMs struggle with long-horizon spatial consistency and relative orientation. 
By enforcing strict directional accuracy over multi-step paths, Spatial4D-Bench effectively disentangles true embodied route planning capabilities from lucky guesses driven by language priors.

\section{Summary of Findings}
\label{our_findings}

By analyzing model performance across our six-category on Spatial4D-Bench, we distill the following critical insights regarding the current state of 4D spatial intelligence.

\begin{enumerate}
  \item{
    \textbf{
    The human-machine performance gap in 4D spatial intelligence remains significant.} Our evaluation reveals a stark divergence between \textit{understanding} and \textit{reasoning}.
    Substantial efforts are required to further improve MLLMs toward human-level spatial cognition.
    \begin{itemize}
      \item{
    \textbf{MLLMs have reached or even surpassed human-level spatial cognition on some perception-related tasks in \textit{object understanding} and \textit{scene understanding}.} 
    These results suggest that perception-related spatial intelligence has been largely solved by cutting-edge models\footnote{For \textit{object attribute estimation}, we elevate the difficulty by injecting significant reasoning component to the QA pairs, making it challenge for MLLMs to infer correctly.}.
  }
  \item{
    \textbf{MLLMs usually perform significantly worse than humans on \textit{spatial reasoning} and \textit{spatiotemporal reasoning} tasks.}
    This highlights a fundamental deficiency of MLLMs in maintaining a coherent 4D world model over extended temporal sequences and in grounding visual perception based on fundamental physical laws.
    }  
    \end{itemize}
  }
    \item{
    \textbf{MLLMs are relatively fragile in spatiotemporal continuity, and long-context temporal modeling remains a bottleneck.}
    MLLMs experience an noticeable performance degradation when dealing with long videos.
    Solving 4D spatial intelligence requires a paradigm shift from fixed-context windows to adaptive sampling or streaming memory architectures.
    }
    \item{
    \textbf{MLLMs exhibit a knowledge-perception gap in intuitive physics.} MLLMs possess abstract knowledge of physical laws but lack the visual grounding to detect the violations in the physical world.
    }

  \item{
    \textbf{MLLMs perform significantly better with multimodal inputs (video and text) than with text-only inputs.}
    This indicates that 4D spatial intelligence requires explicit temporal integration and cannot be solved via language models alone.
  }
  \item{
   \textbf{Language priors can override visual evidence.} 
    Existing MLLMs have a critical limitation that they struggle to weigh conflicting evidence, often letting incomplete visual data override reliable language priors.
  }
  \item{
    \textbf{Egocentric route plan with MLLMs remains an unsolved problem.} 
    Existing MLLMs struggle to construct a coherent spatial map from egocentric videos and rely on hallucination in egocentric route plan, effectively reverting to random guessing or language priors even when visual data are provided.
  }
  \item{\textbf{Open-source models are effectively closing the performance gap.} 
    While proprietary models maintain a lead, top-tier open-source models have achieved comparable performance. The margin between the best proprietary and open-source systems is relatively narrow compared to the gap with human performance.
  }
\end{enumerate}

\section{Conclusion}
\label{conclusion}

In this work, we present Spatial4D-Bench, a large-scale, multi-task 4D spatial intelligence benchmark designed to comprehensively assess the spatial reasoning abilities of MLLMs.
Spatial4D-Bench comprises \textasciitilde40,000 question-answer pairs which are organized into 6 categories covering 18 well-defined tasks that parallel the versatility of human spatial intelligence. 
This significantly distinguishes Spatial4D-Bench from existing benchmarks.
Our thorough experiments on Spatial4D-Bench with 11 state-of-the-art open-source and proprietary MLLMs reveal that MLLMs still exhibit a performance gap relative to humans in comprehensive 4D spatial reasoning.
We have presented various findings derived from extensive experiments conducted on Spatial4D-Bench, which can provide valuable insights to the community.
We hope that the release of  Spatial4D-Bench facilitates the development of more capable MLLMs toward human-level 4D spatial intelligence.

\bibliographystyle{plain}
\bibliography{ref}

\newpage
\appendix

\section*{Appendix}

\begin{table*}[ht]
    \centering
   \scriptsize
    \begin{tabular}{p{0.15\linewidth} | p{0.4\linewidth} | p{0.3\linewidth} | p{0.05\linewidth}  }
      \toprule 
      \textbf{Task}  & \textbf{Example questions} & \textbf{Example answer options} & \textbf{GT}\\
      \hline
      \hline
      Object Size Estimation   & How long is the longest side of the stove, measured in centimeters? & N/A & "62" \\
      \hline
      Object Attributes Estimation   & About the yoga mat on the armchair, which pair is the correct description of its attributes?   & ['A. Oval bases with hole on one side.' 'B. Oval bases and no holes.' 'C. Circular bases with holes on both side.' 'D. Circular bases with hole on one side.'] &  'C' \\
      \hline
      Object Counting   & Count the number of table(s) in the video. & N/A & 2 \\
      \hline
       Affordance Estimation   & Base on the video, if i want to wash hands, dishes, or small items using running water in a fixed basin with drainage, which position allows me to accurately locate the object to do it?   & ['A. It is on the kitchen cabinet.' 'B. It is below the kitchen cabinet.' 'C. It is near the counter.' 'D. It is below the cabinet.'] & 'C' \\
      \hline  
      Room Size Estimation   & Estimate the floor area of the space in square meters, including all visible rooms. & N/A & "26.4" \\
      \hline 
      Scene Classification   & Which of the following descriptions of the scene in the video is the most accurate? & ['A. A living room, zero sofas, zero bookshelfs, and one tv.' 'B. A living room, one sofa, one bookshelf, and one tv.' 'C. A loft, one sofa, one bookshelf, and one tv.' 'D. A living room, fewer sofas than bookshelfs and one tv.'] & "B" \\
      \hline
      3D Grounding & Detect the 3D bounding box of the sofa. Coordinate System Definition: X-axis points rightward, Y-axis points downward, and Z-axis points forward, the origin point is the position of the camera in the first video frame. The format of the answer is [x, y, z, l, w, h, pitch, yaw, roll]. Note: (1) x, y, z: the center of the object in the coordinate system, in centimeters. (2) l, w, h: the dimensions of the object along the XYZ axes, in centimeters, when the rotation angles are zeros. (3) pitch, yaw, roll: Euler angles representing rotations around the X, Y, and Z axes, respectively. Each angle lies between (0, 360). Select the most likely 3D bounding box.    & ['A. [40, 246, -19, 54, 36, 83, 190, 4, 167]' 'B. [272, 216, -9, 66, 25, 91, 184, 349, 260]' 'C. [-158, 45, 10, 92, 166, 84, 176, 10, 82]' 'D. [-106, 279, -32, 42, 46, 85, 171, 352, 332]'] & "C" \\
      \hline
      Absolute Distance Estimation  & What is the shortest distance (in meters) between the sofa and the stove measured from their closest edges? & N/A & 2.9 \\
      \hline
      Relative Distance Estimation   & Considering the closest point on each object, which of chair, stool, stove, sofa is the nearest to the TV? & ['A. chair' 'B. stool' 'C. stove' 'D. sofa'] & "A" \\
      \hline
      Relative Orientation Estimation  & From the perspective of standing at the stove and looking toward the sofa, where is the TV located relative to me: front-left, front-right, back-left, or back-right? & ['A. back-left' 'B. front-right' 'C. front-left' 'D. back-right'] & "C" \\
      \hline
      Action Recognition   & Which of the following is the correct temporal order of these steps?   & [
        "['A. heat some oil add some combined spice and stir -> add some chopped chicken breast and coat it with the mixture -> add some water and cover with a lid -> add some chopped tomatoes and mix it -> season the dish with some chopped green onion' 'B. add some chopped chicken breast and coat it with the mixture -> heat some oil add some combined spice and stir -> add some water and cover with a lid -> add some chopped tomatoes and mix it -> season the dish with some chopped green onion' 'C. heat some oil add some combined spice and stir -> add some chopped chicken breast and coat it with the mixture -> add some chopped tomatoes and mix it -> add some water and cover with a lid -> season the dish with some chopped green onion' 'D. heat some oil add some combined spice and stir -> add some water and cover with a lid -> add some chopped chicken breast and coat it with the mixture -> add some chopped tomatoes and mix it -> season the dish with some chopped green onion'] & 'A' \\
      \hline    
    \bottomrule
    \end{tabular}
    \caption{Example questions and answers (QA) for the $18$ tasks supported in our benchmark, part 1/2.}
    \label{tab:example_qas_table_1}
\end{table*}

\begin{table*}[ht]
    \centering
   \scriptsize
    \begin{tabular}{p{0.15\linewidth} | p{0.4\linewidth} | p{0.3\linewidth} | p{0.05\linewidth}  }
       \toprule 
      \textbf{Task}  & \textbf{Example questions} & \textbf{Example answer options} & \textbf{GT}\\
      \hline
      \hline
      Appearance Order   &  In what sequence do the following categories first appear in the video: towel, door, mirror, basket? & ['A. towel, basket, mirror, door' 'B. towel, basket, door, mirror' 'C. towel, door, mirror, basket' 'D. towel, mirror, door, basket'] &  'A' \\                            
      \hline   
      Spatial Memory   & How did the status of stacked shelf levels change in the video?    & ['A. They have been fully installed in the portable stand' 'B. They never used or touched' 'C. They have been only accidentlay touched but nothing more' 'D. They have been touched, and the top ones also get out for inspection, but still stacked together till the end'] & 'D' \\
      \hline
      State Change Detection  & What happened to the person's shoes? & ['A. The person wore them and walked up, took them off on top of stairs.' 'B. They remained on the person's feet.' 'C. The person took them off at the beginning.' 'D. The person did not wear any shoe in this video.'] & "B" \\
      \hline 
      Egocentric Reasoning  & The picture on the wall is west of the tall floor lamp. Where is the brown three seat sofa positioned relative to the window that is further from table with plant pots on top? & ['A. Northeast' 'B. Northwest' 'C. Southwest' 'D. Southeast'] & "B" \\  
      \hline
      Route Plan  & You are a robot beginning at hallway1 facing the big glass window. You want to navigate to bedroom1. You will perform the following actions (Note: for each [please fill in], choose either A, B, C, D or E): 1. go forward towards the big windows and stop before the kitchen island in the kitchen. 2. turn right. 3. go forward towards the stove. 4. turn left. 5. go towards the door that has a glass beam. 6. [please fill in]. 7. [please fill in]. 8. go towards bedroom1 and stop before the bed. You have reached the final destination. & ['A. turn left' 'B. turn right' 'C. turn 180 degrees' 'D. go upstairs' 'E. go downstairs'] & ["B","A"] \\
      \hline    
      Action Prediction   & What is the person going to do?   & "['A. He will sit down and wait without using the machine.' 'B. He will pour laundry detergent on top of his clothes.' 'C. He will put his card into the laundry machine's slot and press the buttons.' 'D. He will use the blue machine to check his balance.'] & 'C' \\
      \hline  
      Physical Plausibility Reasoning  & Which option violates physical common sense the most in this video?   & ['A. The rider's leg appears fused with the scooter handle.' 'B. The scooter rolls forward while both wheels remain perfectly still.' 'C. The scooter's shadow peels off the ground and climbs the wall.' 'D. The rider and scooter briefly levitate together above the pavement.'] & 'A' \\
    \hline
    \bottomrule
    \end{tabular}
    \caption{Example questions and answers (QA) for the $18$ tasks supported in our benchmark, part 2/2.}
    \label{tab:example_qas_table_2}
\end{table*}

\begin{table*}[ht]
    \centering
   \footnotesize
    \begin{tabular}{p{0.25\linewidth} | p{0.6\linewidth} }
      \toprule 
      \textbf{Task}  & \textbf{Question Template} \\
      \hline 
      \hline
      Object Size Estimation & How long is the longest side of object1, measured in centimeters? \\
      \cdashline{2-2}
      & What is the size of object1 along its longest axis, in centimeters? \\
      \hline
      Object Cunting & Count the number of object1 in the video. \\
      \cdashline{2-2}
      & What’s the number of object1 in the video? \\
      \hline
      Room Size Estimation & Estimate the floor area of the space in square meters, including all visible rooms. \\
      \cdashline{2-2}
      & How many square meters is this space? Include all visible rooms. \\
      \hline
      Scene Classification & Which of the following descriptions of the scene in the video is the most accurate? \\
      \hline
      3D Grounding & Detect the 3D bounding box of object1. Coordinate System Definition: X-axis points rightward, Y-axis points downward, and Z-axis points forward, the origin point is the position of the camera in the first video frame. The format of the answer is [x, y, z, l, w, h, pitch, yaw, roll]. Note: (1) x, y, z: the center of the object in the coordinate system, in centimeters. (2) l, w, h: the dimensions of the object along the XYZ axes, in centimeters, when the rotation angles are zeros. (3) pitch, yaw, roll: Euler angles representing rotations around the X, Y, and Z axes, respectively. Each angle lies between (0, 360). Select the most likely 3D bounding box.\\
      \hline
      Absolute Distance estimation & What is the shortest distance (in meters) between object1 and object2, measured from their closest edges? \\
      \cdashline{2-2}
      & What is the minimum distance between object1 and object2, expressed in meters? \\
      \hline
      Relative Distance Estimation & Considering the closest point on each object, which of obj1, obj2, obj3, or obj4 is the nearest to the obj0? \\
      \cdashline{2-2}
      & Among the listed objects (obj1, obj2, obj3, or obj4), which one is the nearest to the obj0? \\
      \hline
      Relative Orientation Estimation & From the perspective of standing at object1 and looking toward object2, where is object3 located relative to me: front-left, front-right, back-left, or back-right? \\
      \cdashline{2-2}
      & With object1 as my location and object2 as my line of sight, is object3 to my front-left, front-right, back-left, or back-right? \\
      \hline
      Action Recognition & The Video has ** frames at ** FPS. What step is shown between frame ** and frame **? \\
      \cdashline{2-2}
      & Which of the following is the correct temporal order of these steps? \\
      \hline
      Appearance Order Recognition & What will be the first-time appearance order of the following categories in the video: obj1, obj2, obj3, obj4?\\
      \hline
      Egocentric Reasoning & When you took image1/image2, where was the camera for image2/image1, relative to you?\\
      \cdashline{2-2}
      & Which direction is object1 relative to me when I am taking image1/image2? \\
      \cdashline{2-2}
      & When you are taking the last image, in which direction is area1 relative to you? \\
      \cdashline{2-2}
      & Object1 sits/is west of object2. Where is object3 positioned relative to object1/object2/object4? \\
      \cdashline{2-2}
      & In which direction is object1 relative to area1, with object2 on the north/west wall?\\
      \cdashline{2-2}
      & Object1 is east/west/south/north of object2. Where is area1 located relative to area2? \\
      \cdashline{2-2}
      & Based on the continuous image, in which direction is the camera rotating? \\
      \cdashline{2-2}
      & With the camera facing forward to take the two images, assuming a person facing backward, relative to the person, in which direction is object1 moving? \\
      \hline
      Route Plan  & You are a robot beginning at hallway1 facing object1. You want to navigate to area1. You will perform the following actions (Note: for each [please fill in], choose either A, B, C, D or E): 1. go forward towards the big windows and stop before the kitchen island in the kitchen. 2. turn right. 3. go forward towards the stove. 4. turn left. 5. go towards the door that has a glass beam. 6. [please fill in]. 7. [please fill in]. 8. go towards bedroom1 and stop before the bed. You have reached the final destination. \\
    \bottomrule
    \end{tabular}
    \caption{Example question templates.}
    \label{tab:qu_tem1}
\end{table*}

\end{document}